\documentclass[10pt,twocolumn,letterpaper]{article}

\usepackage{cvpr}              

\usepackage{graphicx}
\usepackage{amsmath}
\usepackage{amssymb}
\usepackage{booktabs}
\usepackage{graphicx}
\usepackage{amsmath}
\usepackage[utf8]{inputenc} 
\usepackage[T1]{fontenc}    
\usepackage{url}            
\usepackage{booktabs}       
\usepackage{amsfonts}       
\usepackage{nicefrac}       
\usepackage{microtype}      
\usepackage{xcolor}         
\usepackage{float}  
\usepackage{picinpar}
\usepackage{color}
\usepackage{multirow}
\usepackage{picinpar}
\usepackage{multicol}
\usepackage{colortbl}
\usepackage{caption}

\usepackage[pagebackref,breaklinks,colorlinks]{hyperref}

\usepackage[capitalize]{cleveref}
\crefname{section}{Sec.}{Secs.}
\Crefname{section}{Section}{Sections}
\Crefname{table}{Table}{Tables}
\crefname{table}{Tab.}{Tabs.}

\def\cvprPaperID{3558} 
\def\confName{CVPR}
\def\confYear{2023}
\newcommand\green[1]{\textcolor{green}{#1}}
\newcommand\purple[1]{\textcolor{purple}{#1}}

\newcommand\blue[1]{\textcolor{blue}{#1}}

\newcommand{\red}[1]{\textcolor{red}{#1}}
\definecolor{light-yellow}{rgb}{1,0.768,0.0625}

\begin{document}

\title{Multiview Detection with Cardboard Human Modeling}

\author{Jiahao Ma\textsuperscript{1,2,}\footnotemark[1]\ , Zicheng Duan\textsuperscript{2,}\thanks{The authors contribute equally}\ , Liang Zheng\textsuperscript{2}, Choung Nguyen\textsuperscript{1} \\
CSIRO Data61\textsuperscript{1}, Australian National University\textsuperscript{2}\\
{\tt\small \{jiahao.ma, zicheng.duan, liang.zheng\}@anu.edu.au, choung.nguyen@csiro.au}
}

\maketitle
\begin{figure*}
\setlength{\belowcaptionskip}{-0cm}
  \centering{
    \includegraphics[width=0.99\linewidth]{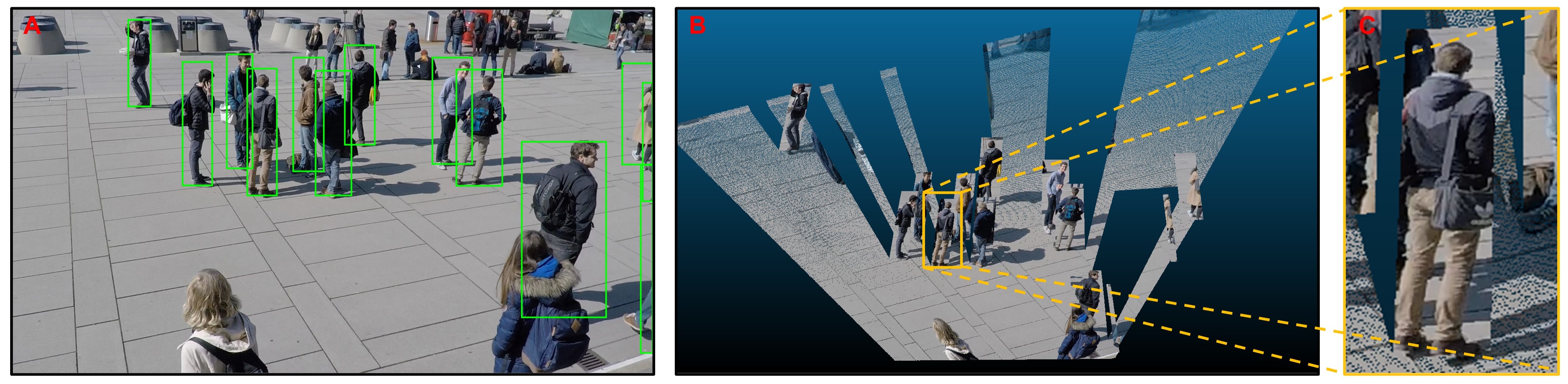}}
  \caption{Illustration of modeling humans as cardboard point clouds. {\bf A}: Our system detects pedestrians in the 2D bounding boxes. {\bf B}: After computing the depth of pedestrians and the ground plane by ray tracing, we project them to the 3D space. {\bf C}: We zoom in on the pedestrian in the {\color{light-yellow} yellow box} in Figure B. In this procedure, cardboard modeling refers to the upright, thin cardboard-like human point clouds of size $w\times h \times 1$, where $w$ and $h$ are the width and height, respectively, and $1$ means 1 channel (a single plane of thickness 1).  
  The cardboard human point clouds reflect the 3D position, height and appearance of each pedestrian and are later aggregated to find human locations. }

  \label{Fig:human_modeling}
\end{figure*}

\begin{abstract}
   Multiview detection uses multiple calibrated cameras with overlapping fields of views to locate occluded pedestrians. In this field, existing methods typically adopt a ``human modeling - aggregation'' strategy. To find robust pedestrian representations, some intuitively incorporate 2D perception results from each frame, while others use entire frame features projected to the ground plane. However, the former does not consider the human appearance and leads to many ambiguities, and the latter suffers from projection errors due to the lack of accurate height of the human torso and head. In this paper, we propose a new pedestrian representation scheme based on human point clouds modeling. Specifically, using ray tracing for holistic human depth estimation, we model pedestrians as upright, thin cardboard point clouds on the ground. Then, we aggregate the point clouds of the pedestrian cardboard across multiple views for a final decision. Compared with existing representations, the proposed method explicitly leverages human appearance and reduces projection errors significantly by relatively accurate height estimation. On four standard evaluation benchmarks, the proposed method achieves very competitive results. Our code and data will be released at \url{https://github.com/ZichengDuan/MvCHM}.

\end{abstract}

\section{Introduction}
\label{sec:intro}

Multiview detection, a.k.a. multi-camera detection, usually refers to detecting objects using images from multiple viewpoints. 
This setup is especially advantageous when the scene is under heavy occlusion, which causes difficulties for monocular detection systems.

Existing methods in this field adopt two general steps: feature extraction and aggregation. The former aims to leverage the scene geometry to extract discriminative  descriptors for pedestrians (and the scene). The latter fuses what is extracted from all the viewpoints and uses a regressor to locate pedestrians on the ground plane. This paper focuses on improving the first step, especially on how to leverage the provided calibration parameters. 

Literature broadly has two strategies to model humans. 
Some intuitively and simply use 2D perception results such as 2D bounding boxes \cite{Lima_2021_CVPR, xu2016multi, deepocclusion}, segmentation/foreground pixels \cite{PR, pomcnn} to represent individual pedestrians, which are later clustered on the ground plane. While these methods have strong generalization ability and are interpretable, they merely use mathematical geometry relations to cluster pedestrians' positions without considering their \textit{appearance} feature, typically leading to inaccurate aggregation outcomes. Others leverage camera calibration to project features of \textit{entire image frames} onto the ground plane, which are used to collectively represent pedestrians \cite{mvdet, shot, mvdetr, vfa, 3DROM}. Compared with the first strategy, these approaches use both pedestrian location and appearance features, aggregating frame features to obtain improved performance. However, they suffer from inaccurate feature projections: there lacks an estimation of the pedestrian height, so pixels along the same vertical line in the 3D coordinate system are not projected onto the same point on the ground plane. 

Considering the above discussions, we introduce a new multiview detection method with cardboard human modeling referred to as MvCHM. In a nutshell, we first detect pedestrians in each camera view in a plain way, and then build human point clouds, which are defined as cardboard human modeling, after estimating the depth of the standing point (the location where a pedestrian stands) and head. The point clouds from all the views are fed into a neural network for aggregation and location regression. This pipeline is illustrated in Fig.  \ref{Structure}, where an interesting component is the cardboard-like human point clouds made up of only one channel of pixels, as shown in Fig.\ref{Fig:human_modeling}. Compared with existing works, they contain more accurate human appearance and location to be further vectorized by the neural network. 


Our method has a few advantages. First, compared with the ``2D perception + clustering'' methods \cite{deepocclusion, PR, Lima_2021_CVPR, xu2016multi, pomcnn}, we incorporate sufficient human appearance features into this pipeline. Second, compared with the ``projection + aggregation'' methods \cite{mvdet, shot, mvdetr, vfa, 3DROM}, our method significantly reduces projection errors caused by inaccurate height through the cardboard modeling process and thus provides more accurate human appearance features.

We evaluate our system on four multiview pedestrian detection benchmarks WildTrack \cite{wildtrack}, MultiviewX \cite{mvdet} and their extension. We show that the proposed point clouds processing procedures enabled by the aggregation network give very competitive results.
\section{Related work}
\subsection{Feature-projection based multiview detection}
Generally, feature-projection-based methods \cite{mvdet, shot, mvdetr, vfa, 3DROM} project multiview high-resolution feature maps to the ground plane, concatenate these features and regress object positions from the features. Hou \textit{et al}. \cite{mvdet} project convolution feature maps to the ground plane via a perspective transformation and adopt a full convolution network to aggregate the concatenated feature maps. Motivated by \cite{mvdet}, Song \textit{et al}. \cite{shot} introduce stacked homography transformation to project frame features to the ground plane at different height levels. Hou \textit{et al}. \cite{mvdetr} deal with shadow-like distortions in different cameras and positions via a transformer structure. To align features along the vertical direction of objects, Ma \textit{et al}. \cite{vfa} voxelize 3D features before multiview aggregation. As mentioned above, projection-based methods are not 3D aware, so would likely encode noisy image content (\emph{e.g.}, background and misaligned human) in the project features. 

\subsection{2D-Perception based multiview detection}
The other line of methods \cite{PR, Lima_2021_CVPR, xu2016multi, pomcnn, deepocclusion} intuitively utilize 2D perception results to model each pedestrian, which are clustered on the ground plane. We call them 2D-perception-based methods. For example, Lima \textit{et al}.  \cite{Lima_2021_CVPR} forfeit training and instead estimate the standing point within each 2D detection bounding box and predict the 3D coordinate of pedestrians by solving the clique cover problem. Yan \textit{et al}. \cite{PR} calculate the likelihood of pedestrian presence in each foreground region and clusters pedestrian positions via minimizing a logic function. Fleuret \textit{et al}. \cite{pomcnn} estimate the probabilities of pedestrian occupancy via a probabilistic occupancy map. To aggregate multiview detection results, mean-field inference \cite{deepocclusion, pomcnn} and conditional random field (CRF) \cite{deepocclusion, CRF} can be exploited. Our work also starts from using 2D perception results, {{\textit{i.e.,} 2D bounding boxes}}, but differs from existing works in that we explicitly consider the human appearance and use regression to find human locations (similar to the projection-based methods).


\subsection{Estimating point clouds in 3D object detection}
In 3D object detection, some existing methods generate scene point clouds by estimating the depth of the scene. Because the point clouds are not provided by LiDAR, they are often called pseudo LiDAR point clouds. 
Wang \textit{et al}. \cite{PseudoLidar} show that a key to closing the gap between image- and LiDAR-based 3D object detection may simply be 3D representations. 
MF3D \cite{MF3D} estimates disparity maps to obtain pseudo LiDAR and fuses input RGB images with front-view features obtained by the disparity map. Mono3D-PLiDAR \cite{Mono3D_PLiDAR}, a two-stage 3D object detection pipeline, converts input images into point clouds via DORN \cite{DORN} and applies Frustum PointNets \cite{FrustumPointNet} to localize 3D objects. While these works use end-to-end pixel-wise depth estimation methods, we calculate the depth value of detected pedestrian regions via the \emph{ray tracing} technique given camera poses, a ground plane, and pedestrians standing points on the ground.

\begin{figure*}
\setlength{\abovecaptionskip}{-0cm}
  \centering{
    \includegraphics[width=0.98\linewidth]{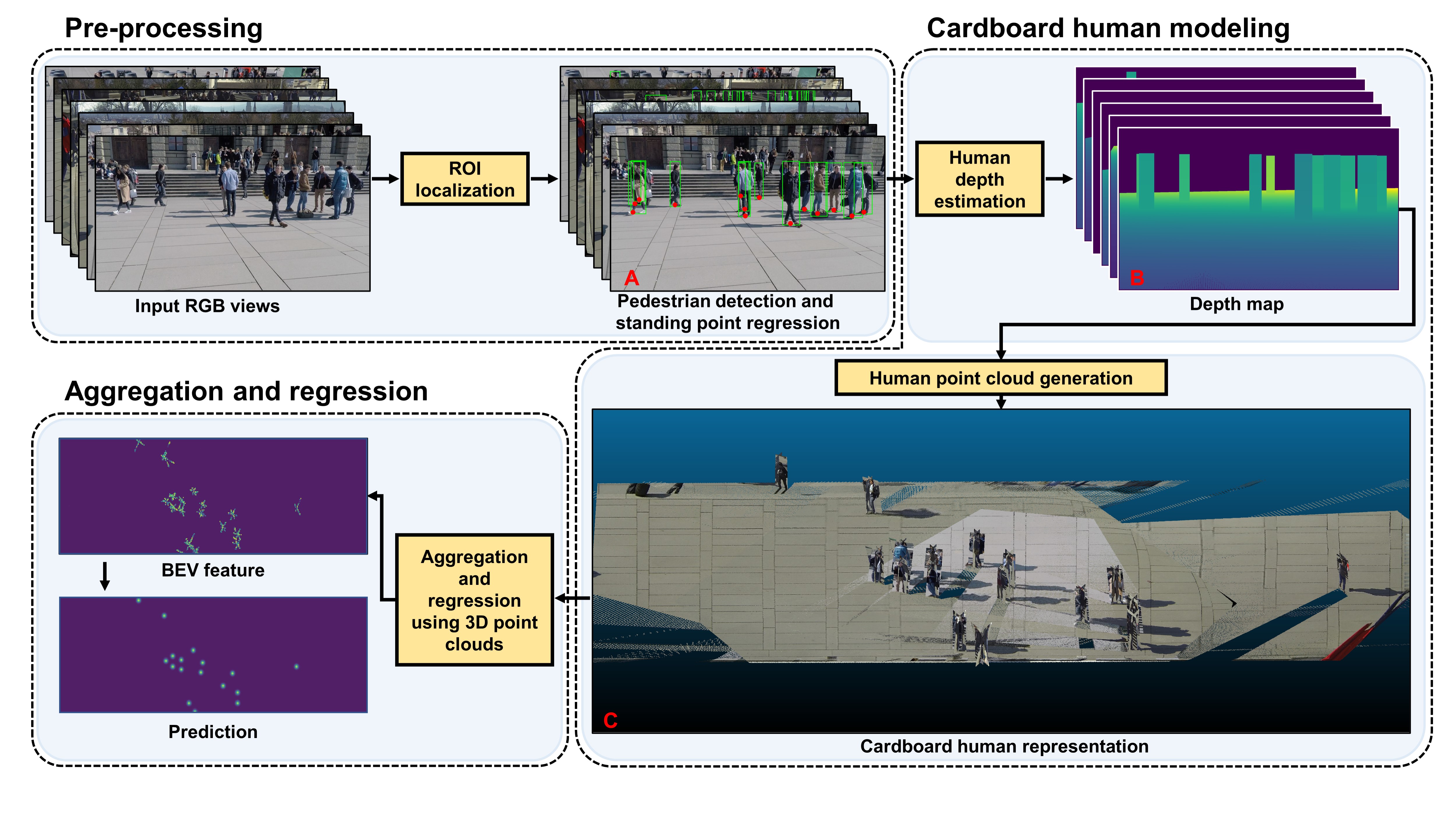}}
  \caption{The proposed system pipeline. 
  First, given input RGB images from each view, we apply 2D object detection to obtain per-view pedestrian detection results. Then, through keypoint detection, we find the \textit{standing point} of each detected bounding box. Next, we estimate the depth of the standing point and head of each detected person and 
  fill the whole body region with the interpolated depth. With the estimated depth we project the detection results into the 3D space and generate human cardboard point clouds (with only one channel). Finally, we use an aggregation and regression network to 
  find the occupancy heat map on the bird's-eye-view (BEV) plane. }
  \label{Structure}
\end{figure*}

\section{Preliminaries: Ray tracing to compute 3D coordinates of a 2D point}
\label{sec:ray-tracing}
Ray tracing is a technique for modeling light transport: a light ray emerges at the light source, reflects on objects, and goes into the camera.
Formally, ray tracing is formulated as:
\begin{equation}
\label{ray-tracing-formula}
P = O + tD.
\end{equation}
This formula computes the \textit{3D coordinates of a reflection point} on an object, denoted as $P=\begin{bmatrix}
  P_x, P_y, P_z
\end{bmatrix}^T$. 
$O=[O_x, O_y, O_z]^T$ is the 3D position of the camera, or origin; $D = [D_x, D_y, D_z]^T$ is the direction of the ray; $t\in \mathbb{R} $ is the distance between the camera and the reflection point on the object. 
where $O$ and $D$ are accessible with camera pose. Using Eq. \ref{ray-tracing-formula}, we compute the 3D coordinates and thus the depth of the standing point and the head of each pedestrian. 

\section{Proposed system}
As shown in Fig.  \ref{Structure}, our system consists of pre-processing (Section \ref{sec:pre-processing}), human modeling (Section \ref{sec:human_modeling}) and aggregation (Section \ref{sec:aggregation}), where \textit{human modeling is our main contribution}. Below we will detail these steps with a focus on the human modeling process, including human depth estimation and human point clouds generation. 

\subsection{Pre-processing}
\label{sec:pre-processing}
Pre-processing, also denoted as ROI localization in Fig. \ref{Structure} A, aims to find 1) pedestrian regions in the shape of bounding boxes and 2) the standing point of each person. Both will be used in Section \ref{sec:human_modeling} for depth estimation and person height calculation. 
We firstly adopt a 2D detector CrowdDet \cite{CrowdDet} to detect bounding boxes. Regarding the latter, following \cite{Hourglass, MSPN}, we use a regression neural network to obtain the positions on the ground where pedestrians stand. Essentially, the detected bounding boxes are used as input, and the output is a single standing point. In the implementation, we use the standing point annotations provided by the benchmarks and mainly use \cite{MSPN} for regression, with a comparison with \cite{Hourglass}.

\subsection{From 2D to 3D: Cardboard human modeling}
\label{sec:human_modeling}

In this section, we describe the proposed cardboard modeling that transforms 2D bounding boxes into 3D point clouds shaped as standing cardboard on the ground plane.
Specifically, based on the located ROI, a human in a 2D image is modeled as a cardboard-like point clouds of size $w\times h \times 1$ (refer Fig. \ref{Fig:human_modeling}) in the 3D space. 
These point clouds reflect a pedestrian's appearance, height and 3D spatial position, and will be used for human feature extraction and localization (Section \ref{sec:aggregation}). 
Generating the cardboard human is simple: we calculate the depth of each pedestrian using ray tracing, and then project the pedestrian into the 3D sapce.

\textbf{Human depth estimation.}
Due to the lack of pixel-wise human depth annotations, it is infeasible to estimate accurate depth for each human pixel. To get around this problem, we compute the depth of the standing point and the head using the ray tracing technique \cite{ray-tracing} (Section \ref{sec:ray-tracing}). The two depth values are subsequently used to interpolate the depth of other pixels in the bounding box in a linear way.

\begin{figure}
\setlength{\belowcaptionskip}{-0cm}
  \centering{
    \includegraphics[width=0.99\linewidth]{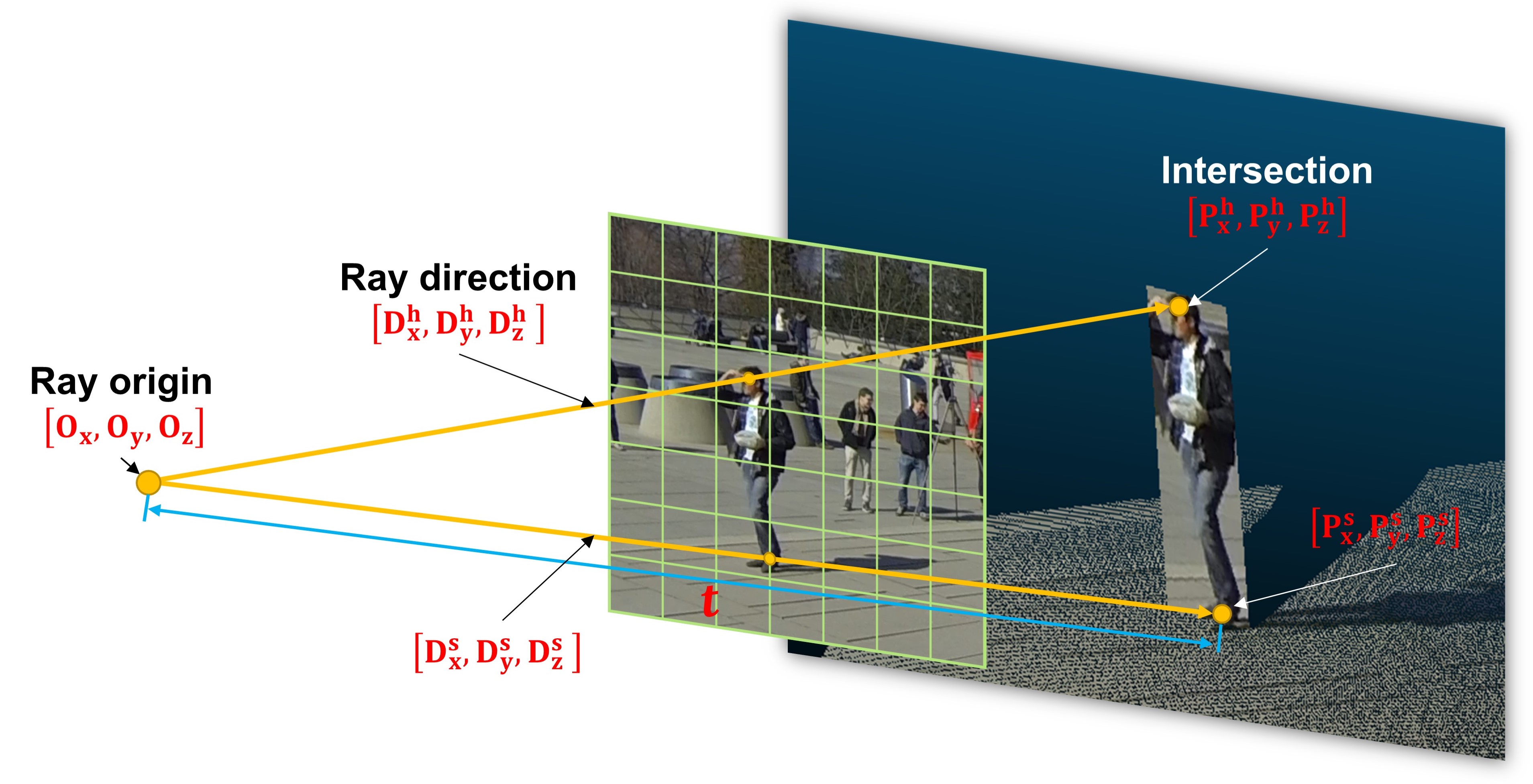}}
  \caption{Illustration of ray tracing. According to the reversibility of the ray, we define the camera center (ray origin) as $O$, the ray direction as $D$, the object 3D location as $P$, and the distance between the camera and reflection points on the object as $t$. First, we assume the pedestrian standing point is on the ground surface where $P_z = 0$, then given the camera matrix and the 2D coordinates of the standing point in the image, the depth of the standing point can be accurately calculated, the depth of the head is further calculated by substitution, finally, the depth of the rest of body region is linearly interpolated. Detailed derivations are presented in supplementary materials \ref{ray-tracing-derivation}.}
  \label{ray_tracing}
\end{figure}

We now leverage Eq. \ref{ray-tracing-formula} to find the 3D coordinates of the standing point and the head, denoted as $P_\text{standpoint} = [P_\text{x}^\text{s}, P_\text{y}^\text{s}, P_\text{z}^\text{s}]$ and $P_\text{head} = [P_\text{x}^\text{h}, P_\text{y}^\text{h}, P_\text{z}^\text{h}]$, respectively, as shown in Fig \ref{ray_tracing}. On the one hand, to compute $P_\text{standpoint}$, we assume that all pedestrians are standing on the ground plane with $P_\text{z}^\text{s}=0$. This assumption intuitively holds in normal scenarios. 
Based on the ground plane assumption above, we just need to calculate $O$ and $D$ to get $P_\text{standpoint}$. We first use the provided camera pose to obtain the camera 3D position $O$, then we calculate the ray direction $D$ through the standing point with the image coordinates of the standing point and camera intrinsic, and then obtain $P_\text{x}^\text{s}$ and $P_\text{y}^\text{s}$ by substituting $O$ and $D$ into Eq. \ref{ray-tracing-formula}. Next, to find the pedestrian head in a 3D coordinates system, we assume the standing point and head of each pedestrian have the same x and y coordinates, \emph{i.e.,} 
$P_\text{x}^\text{s}=P_\text{x}^\text{h}$, $P_\text{y}^\text{s}=P_\text{y}^\text{h}$. Similar to calculating the $z$ coordinate of the standing point $P_\text{standpoint}$, we solve the $z$ coordinate of the head $P_\text{z}^\text{h}$ by substituting $P_\text{x}^\text{h}$ or $P_\text{y}^\text{h}$ into Eq. \ref{ray-tracing-formula}. 
After computing the 3D coordinates of the standing point and head in the world coordinate system and then converting them into the camera coordinate system, we default the value of the $Z$ of the camera coordinate to the depth value.

Finally, we use linear interpolation to fill the rest of the detection region with a rough depth value, the generated depth map is shown in Fig. \ref{Structure}B. More derivation details are provided in the \ref{ray-tracing-derivation} of supplementary material.

\label{sec:point_cloud}
{\bf Point clouds generation.} 
After assigning each pixel in the pedestrian region with a depth value, 
we project the pedestrian region from the 2D to 3D space as point clouds according to the intrinsic and extrinsic parameters and estimated depth. Projection details are shown in Section \ref{ray-tracing-derivation} of supplementary materials.
Our experiments in Section \ref{groundplane_samplerate_impact} show that the ground plane point clouds introduce noisy features and additional computational cost, leading to poorer model performance. Therefore, we only project the pedestrian region to the 3D space.

\subsection{Aggregation and regression using 3D point clouds}
\label{sec:aggregation}
Aiming to aggregate features from multiple views, we process point clouds into feature vectors, using the network proposed in \cite{pointpillars}; features are then concatenated to regress pedestrian position on the ground plane. 
Specifically, we discretize point clouds into an evenly spaced grid in the BEV plane, creating a set of pillars (voxels with ultimate spatial extend in the \textit{Z} direction \cite{pointpillars}). Then, we randomly sample the point clouds in each pillar and adopt PointNet \cite{pointnet} to extract high-dimensional features (pillar feature) in each pillar. Based on the pillar representations, we follow \cite{voxelnet} to flatten pillar features to the BEV plane and regress the final pedestrian position. 
Similar to \cite{gaussian_encode, mvdet}, we represent binary ground truth pedestrian occupancy as Gaussian distribution maps. We use the focal loss \cite{focal_loss} as position regression loss:
\begin{equation}
    \mathcal{L}_{reg} = -\alpha (1 - p)^\gamma log_p \ \text{,} \label{focal_loss_eq}
\end{equation}
where $\alpha$ and $\gamma$ are two hyper-parameters. We use the same values of $\alpha$ and $\gamma$ as \cite{focal_loss}.

\begin{figure*}
  \centering{
    \includegraphics[width=0.9\linewidth]{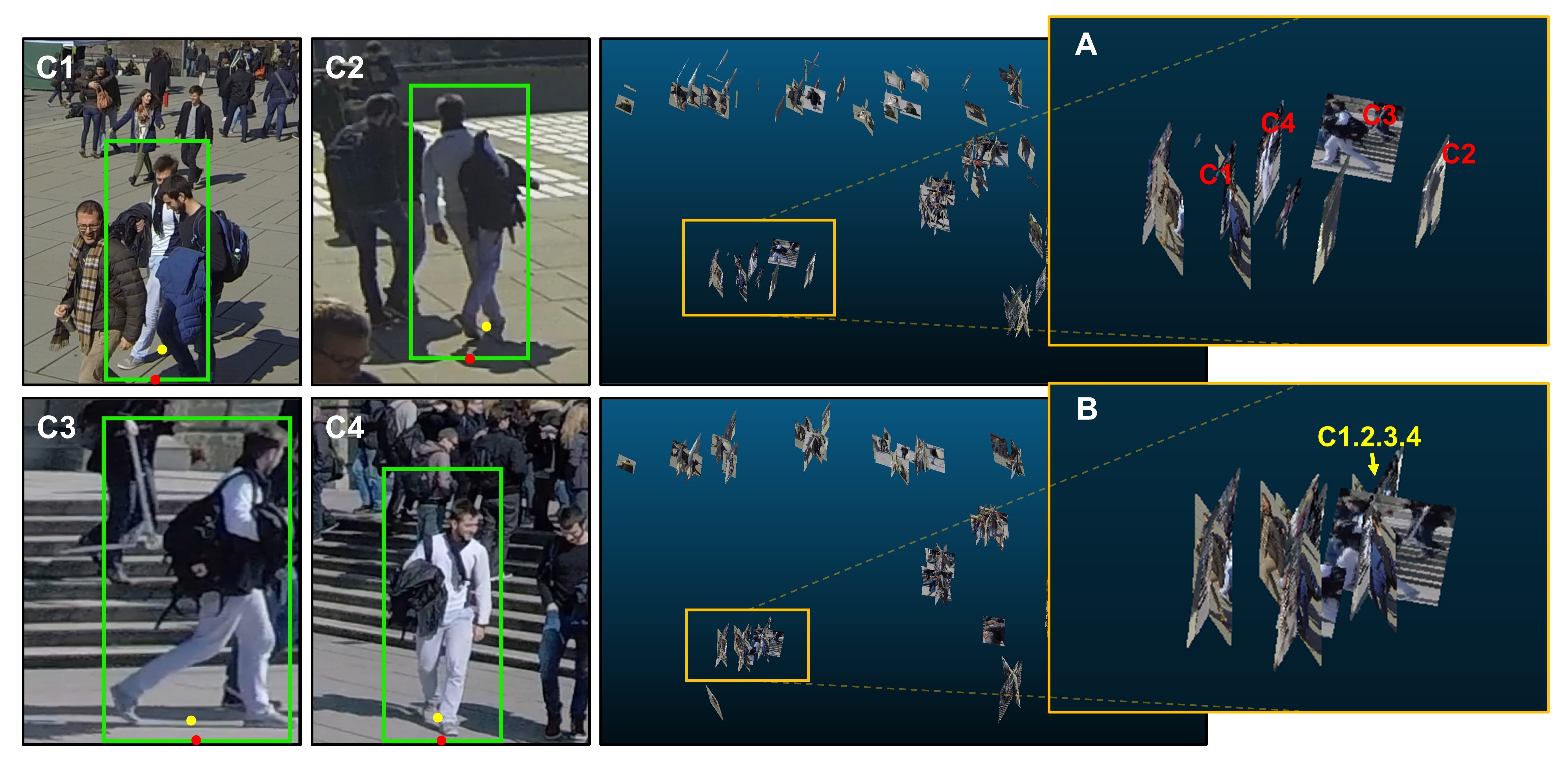}}
  \caption{An illustration of the benefit of standing point estimation. Figures C1 \--{} C4 record four different views capturing the same person. In each view, the green box is the 2D detection result of the person, the red dot is the bottom center of the detected box, and the yellow dot represents the estimated standing point using \cite{MSPN}. Figures A and B show the projection results when regarding the \red{bottom center} or the {\color{light-yellow}estimated point} as the standing point. In figure A and B, each human cardboard is marked with the corresponding view number. We observe that the projected cardboards form denser clusters in figure B, which validates the effectiveness of standing point estimation.} 
  
  
\label{FeetRefine}
\end{figure*}

\section{Discussion}
\textbf{Ours \textit{vs.} projection-based methods: less susceptible to projection noise.} 
As mentioned in previous sections, feature projection-based methods \cite{mvdet, shot, mvdetr, vfa} suffer from inaccurate projections due to wrong human height estimation. In the latter case, part of the pedestrian torso is wrongly projected on the planes with an inaccurate height, and the projected pedestrian features are intermingled with background features or noise.
Our method estimates an accurate height using bounding boxes before projecting the pedestrian region 
to the corresponding 3D space. Thereby, the proposed method effectively recovers the pixel's 3D position along the $Z$ axis and separates human features from the background, which alleviates projection noise.


\textbf{Ours vs clustering-based methods: can take advantage of human appearance.}
By extracting features from the human point clouds, we seamlessly integrate human appearance features into the system. In comparison, existing 2D-perception and clustering-based methods \cite{deepocclusion, PR, Lima_2021_CVPR, xu2016multi, pomcnn} merely use pedestrian 2D position features to predict target position. The drawback of not using human appearance features is experimentally analyzed in Section \ref{variant_study: AppearanceFeature}. 


{\bf Two major performance influencers.} 
Firstly, our system effectiveness relies on 2D detection performance, demonstrated by experiments using various detectors in Fig. \ref{DetectionBackbones}. We show that using a strong 2D detector generally leads to higher accuracy. 
Second, the standing point estimation module also has an impact on the system accuracy. Its impact could be significant because 
the standing point is designated as the projection anchor point for each detected pedestrian, and its localization directly determines the 2D occupancy position of each pedestrian on the ground plane. 
Quantitative analysis of its influence will be shown in Section \ref{variant_study: Pedestrian_KeypointDetection}. Our system, while benefiting from a strong 2D detector and standing point estimator, also benefits from point clouds cardboard modeling (to be shown in Section \ref{variant_study: Pedestrian_KeypointDetection}). All these elements are seamlessly integrated.

\begin{table*}[t!]
  \begin{center}
      \scriptsize
      \resizebox{0.97\linewidth}{!}{
      \begin{tabular}{l cccc c cccc}
      \toprule
      \multirow{2}{*}{Method} & \multicolumn{4}{c}{\bf Wildtrack$^*$} & \quad & \multicolumn{4}{c}{\bf MultiviewX}\\
      \cmidrule[0.3pt]{2-5} \cmidrule[0.3pt]{7-10}
      & MODA & MODP & Precision & Recall & \quad & MODA & MODP & Precision & Recall\\
      \midrule[0.4pt]
      RCNN  \&  clustering\cite{xu2016multi}     &  $11.9^{\S}$ &  $18.1^{\S}$  &  $66.1^{\S}$  &  $44.9^{\S}$  & \quad & $18.7$  & $46.4$  & $63.5$  & $43.9$   \\
      Deep-Occlusion \cite{deepocclusion}   &  -  &  -  &  -  &  -  & \quad &  $75.2$  &  $54.7$  &  $97.8$  &  $80.2$   \\      
      MVDet \cite{mvdet}   &  $88.7$   &  $73.6$   &  $93.2 $  & \green{ $95.4$} & \quad &  $83.9$  &  $79.6$  &  $96.8$  &  $86.7$  \\
      SHOT \cite{shot}  &  $90.8$   & \green{$77.7$} & {$96.0$} &  $94.3$   & \quad &  $88.3$  &  $82.0$  &  $96.6$  &  $91.5$  \\
      MVDeTr \cite{mvdetr}  & \green{$92.1$} & \blue{$84.1$} & \green{$96.1$} &  $94.5$  & \quad & \green{$93.7$} & \red{$91.3$} & \red{$99.5$} & \green{$94.2$}  \\
      3DROM \cite{3DROM} & \blue{$93.9$}  &  $76.0$   & \blue{$97.7$}& \blue{$96.2$} &\quad & \red{$95.0$} & \green{$84.9$} & \blue{$99.0$} & \red{$96.1$}\\
      \bf{MvCHM (ours)} & \red{$95.3$} & \red{$84.5$} & \red{$98.2$} & \red{$97.1$} & \quad & \blue{$93.9$} & \blue{$88.3$} & \green{$98.5$} & \blue{$94.8$} \\

      \cmidrule[0.3pt]{2-5} \cmidrule[0.3pt]{7-10}
      \multirow{2}{*}{} & \multicolumn{4}{c}{\bf Wildtrack+$^*$} & \quad & \multicolumn{4}{c}{\bf MultiviewX+}\\
      
      \midrule[0.4pt]
      RCNN  \&  clustering\cite{xu2016multi}     & 10.1$^{\S}$  &  17.2$^{\S}$  &  65.1$^{\S}$  &  42.3$^{\S}$ & \quad &  19.9$^{\S}$  &  48.9$^{\S}$  &  64.1$^{\S}$  &  44.0$^{\S}$  \\
      Deep-Occlusion   \cite{deepocclusion}    & - & - & - & - & \quad & - & - & - & - \\      
      MVDet \cite{mvdet}   &  $87.8$  &  $74.9$  &  $95.1$  &  $90.7$  & \quad &  $84.5$   &  $80.9$  &  $96.4$  &  $85.2$  \\
      SHOT  \cite{shot}   &  $90.2$  & \green{$77.5$} &  $95.7$  &\green{ $94.1$}  &  \quad &  \green{$88.5$}  &  $82.7$  &  $97.1$  &  $90.2$  \\
      MVDeTr  \cite{mvdetr}  & \green{$92.2$} & \blue{$84.2$} & \blue{$96.3$} &  $94.1$  & \quad & \blue{$93.8$} & \red{$91.5$} & \red{$99.6$} & \green{$93.9$} \\
      3DROM \cite{3DROM} & \blue{$93.8$} &  $77.1$  & \green{$96.9$} & \blue{$96.1$} & \quad & \red{$95.2$} & \green{$85.1$} & \blue{$99.2$} & \red{ $96.7$ }\\
      \bf{MvCHM (ours)} &  \red{$94.6$} & \red{$84.7$} & \red{$98.3$} & \red{$96.6$} & \quad & \blue{$93.8$} & \blue{$87.9$} & \green{$98.6$} & \blue{$95.3$}\\
      \bottomrule
      \end{tabular}
      }
  \end{center}
  \setlength{\abovecaptionskip}{0cm} 
  \setlength{\belowcaptionskip}{-0cm}
  \caption{Comparison with the state-of-the-art methods on the standard evaluation benchmarks. For each metric, the best, second best and third best numbers (in percentage) are highlighted in \red{red}, \blue{blue} and \green{green}, respectively. Our method yields state-of-the-art performances on the Wildtrack/Wildtrack+ datasets and very competitive results on the MultiviewX/MultiviewX+ datasets. On the Wildtrack and MultiviewX datasets, due to the lack of pedestrian training labels outside the detection area, we adapt a pretrained 2D detector in the ROI localization procedure mentioned in Section \ref{sec:pre-processing}, while on the Wildtrack+ and MultiviewX+ datasets, we train a 2D detector using the proposed complete annotations, all other methods follow the same training scheme for fair comparisons. {\bf *} denotes that we use a mask to reduce the effect of the inaccurate labeling, details are discussed in the evaluation Section \ref{sota}, and more visualization on the missing labels are shown in \ref{Label_Ommision} in supplementary materials. ${\S}$ indicates the results are from our implementation.}
  
\label{PerformanceComparison}
\end{table*}

\subsection{Experimental settings}
{\bf Dataset.} We compare our method on two standard multiview pedestrian benchmarks \cite{wildtrack, mvdet}, and two newly created datasets Wildtrack+ and MultiviewX+. 

\textit{Wildtrack} \cite{wildtrack} is a real-world multiview pedestrian detection benchmark capturing people on a square of 12 $\times$ 36 meters with 7 calibrated cameras. The image resolution is 1080 $\times$ 1920, and the square is discretized to a 480 $\times$ 1440 grid. The dataset contains 400 images, the first 360 frames for training and the last 40 for testing.

\textit{Wildtrack+} is an extension of the Wildtrack\cite{wildtrack} dataset, in which we additionally annotate the unlabelled pedestrians outside the detection area. Note that labels inside the detection area remain unchanged. The new annotations allow us to train a 2D detector on Wildtrack instead of borrowing an off-the-shelf detector trained on other datasets.


\textit{MultiviewX} \cite{mvdet} is a synthetic dataset created by Unity for pedestrian detection in crowded scenes. This dataset covers an area of 16 $\times$ 23 meters with 6 synchronized cameras. The ground plane is quantized into a 640 $\times$ 1000 grid, and the resolution is 1080 $\times$ 1920. It also has 400 frames with the last 40 frames for testing.

\textit{MultiviewX+} is newly generated using the same Unity engine following the same labeling mechanism as MultiviewX\cite{mvdet}. Compared with the origin MultiviewX dataset, our MultiviewX+ dataset \textit{1)} additionally annotates the pedestrians outside the detection area to train 2D detectors locally \textit{2)} introduces new character different from that in MultiviewX \textit{3)} provides more accurate camera calibration files.

{\bf Evaluation metrics.} Four metrics are used:  Multiple Object Detection Accuracy (MODA), Multiple Object Detection Precision (MODP), Precision, and Recall. Specifically, MODA accounts for the normalized missed detections and false positives and MODP assesses the localization precision. We estimate the empirical precision and recall, calculated by $P=TP/(TP+FP)$ and $R=TP/(TP+FN)$ respectively. We view MODA as the primary indicator. A threshold of 0.5 meters is used to decide true positives.

For evaluating detection models on the Wildtrack and Wildtrack+ datasets, we observe severe annotation missing near the border of the detection area, leading to an accuracy drop for existing methods. To reduce the impact of missing labels, we mask the border area on both regressed and ground truth heatmap during evaluation, and as a result, all the compared methods now have higher accuracy. More details of the mask are provided in the supplementary materials \ref{Label_Ommision}.

\subsection{Implementation details}
We train the pedestrian detector on the Widltrack+ dataset and MultiviewX+ dataset while borrowing the best-trained model provided by CrowdDet\cite{CrowdDet} on the Wildtrack dataset and the MultiviewX dataset. In training the pedestrian detector, we use the Earth Mover’s Distance loss (EMD Loss) and Set NMS \cite{CrowdDet} which are shown to improve robustness against occlusions. 
For standing point estimation,
we apply the MSPN \cite{MSPN} network and train it with the provided standing point ground truths provided in all four datasets.
When constructing human point clouds, to avoid projection noise, we directly remove the background and merely project pixels in each bounding box to the 3D space.
To train the aggregation and regression network, we use an Adam optimizer with L2 regularization of $5 \times 10^{-3}$. $\alpha$ and $\gamma$ in Eq. \ref{focal_loss_eq} are set to be 2 and 4, respectively. The learning rate is set to $ 2 \times 10^{-4} $. During the evaluation, the heatmap thresholds are set to be 0.8, 0.86, 0.8, 0.8 on the Wildtrack, Wildtrack+, MultiviewX, MultiviewX+ datasets respectively.  We conduct all experiments on a single RTX-3090 Ti GPU. 

\subsection{Evaluation}
\label{sota}
{\bf Comparison with the state-of-the-art methods.} Table \ref{PerformanceComparison} summarizes this comparison. On the Wildtrack dataset, our pipeline achieves state-of-the-art performance: MODA=95.3\%, MODP = 84.5\%, Precision = 98.2\%, and Recall = 97.1\%. In terms of MODA, our method is 1.4\% higher than the second best method 3DROM \cite{3DROM} based on feature projection. On the Wildtrack+ dataset, our approach outperforms other methods with similar margins.

Regarding the MultiviewX and MultiviewX+ dataset, our method is slightly outperformed by the previous state-of-the-art method \cite{3DROM} but still remains very competitive. 
The main reason is that the camera positions in the MultiviewX/MultiviewX+ are lower than that in the Wildtrack dataset, which causes the cameras to look in a relatively horizontal direction, making it difficult to capture the pedestrians' feet. 
The detection results on the evaluation benchmarks are visualized in Fig. \ref{DetectionResults} in the supplementary materials.

\begin{figure}
  \centering{
    \includegraphics[width=0.98\linewidth]{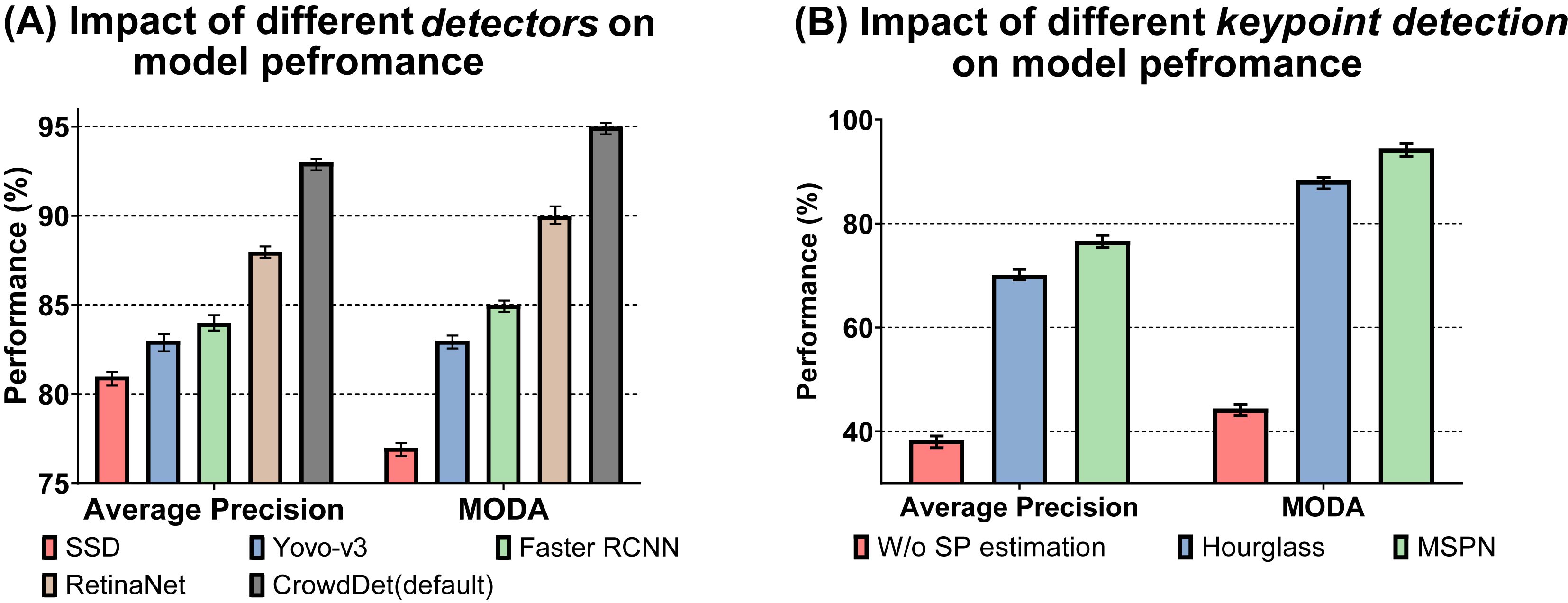}}
\caption{(A): Comparing various pedestrian detectors of their 2D detection accuracy and overall system performance.
(B): Ablation study on having keypoint estimation modules. Verify the impact of standing point detection accuracy on the overall system performance. The results are reported on the Wildtrack dataset.}
  \label{DetectionBackbones}
\end{figure}

\textbf{Necessity of estimating the standing point.}
\label{FeetRefinementAblation}
We perform an ablation study on this module in Fig. \ref{DetectionBackbones}B. For convenient purposes, the standing point is denoted as \textit{SP} in Fig. \ref{DetectionBackbones}B.
During the Pre-processing process introduced in Section \ref{sec:pre-processing}, the standing point estimator (MSPN) regresses the standing point of each person (yellow dot in Fig. \ref{FeetRefine} C1$ \sim $4), where ``$W/o\ \mbox{\textit{SP} estimation}$'' indicates directly regarding the bottom center of the pedestrian bounding box as the standing point (red dot in Fig. \ref{FeetRefine} C1$ \sim $4).
From Fig. \ref{DetectionBackbones}B and \cref{Ablation}, we observe that without the standing point estimation step, system accuracy drops significantly from 95.3\% to 42.1\%. A probable reason for this drop is that the bottom centers of detection bounding boxes usually do not stably indicate the human position (refer to the comparison in A and B of Fig.  \ref{FeetRefine} for the scattered centers).

\begin{figure}
\setlength{\belowcaptionskip}{-0cm}
  \centering{
    \includegraphics[width=1.0\linewidth]{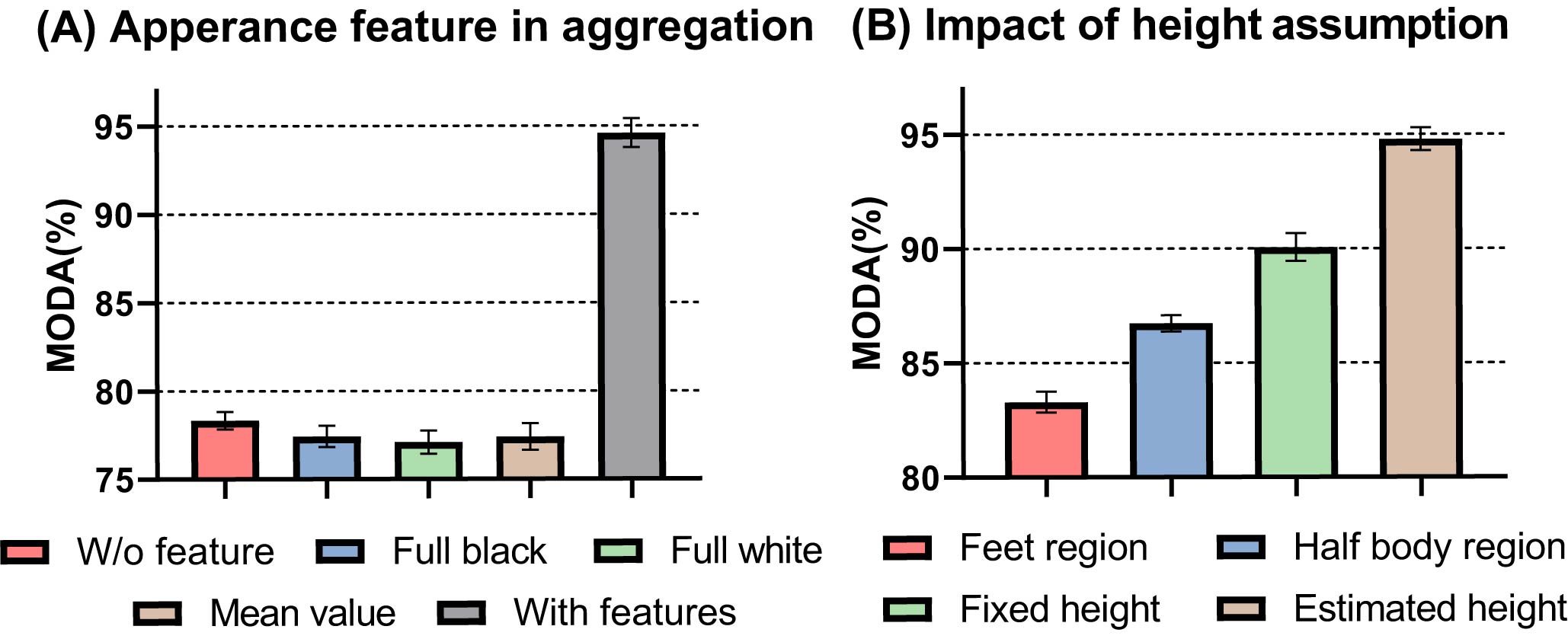}}
\caption{(A): Ablation study on integrating human appearance feature.
(B): Comparing our method with variants in human height estimation.
} 

  \label{HumanFeatureHeight}
\end{figure}

\phantomsection
\label{variant_study: AppearanceFeature}
{\bf Importance of having human appearance features.} As mentioned before, 2D detection-based methods undesirably discard human appearance features, which is unavoidable due to their method designs \cite{PR, deepocclusion, Lima_2021_CVPR, xu2016multi, pomcnn}. In Fig. \ref{HumanFeatureHeight}, we conduct ablation studies to verify the importance of integrating human appearance features. 
In this figure, \textit{w/o feature} means we directly remove RGB from normal point clouds, \textit{i.e.}, changing each point from $[x, y, z, r, g, b]$ into $[x, y, z]$. \textit{``full black''}, \textit{``full white''} and \textit{``mean value''} replace the RGB pixels on the human with black pixels, white pixels and the mean RGB value, respectively. Therefore, these four variants of our method merely encode human location.  
Ablation results in Fig. \ref{HumanFeatureHeight}A and \cref{Ablation} indicate the importance of having human appearance features and these results further validate our design since our method exceeds other variants with a clear margin.

\begin{table}
\setlength{\abovecaptionskip}{0cm}  
  \begin{center}
      \small
      \begin{tabular}{c || c c c || c}
      \hline
      Method & Detector & SP & AF & MODA\\
      \hline
      \multirow{4}*{MvCHM} & \checkmark & & & $20.4$\\
      & \checkmark & & \checkmark & $42.1$\\
      & \checkmark & \checkmark & & $78.1$\\
      & \checkmark & \checkmark & \checkmark & \textbf{$95.3$}\\
      \hline
      \end{tabular}
  \end{center}
  \caption{Modular ablation study reported on the Wildtrack dataset. {\bf SP}: Standing Point detection, {\bf AF}: Appearance Feature.}
\label{Ablation}
\end{table}

\textbf{Comparison of various pedestrian detectors and keypoint detectors.} 2D human detection and standing point detection are two important components of our system. In Fig. \ref{DetectionBackbones}A, we compare CrowdDet \cite{CrowdDet} used in our system with SSD \cite{ssd}, YOLO-v3 \cite{yolov3}, Faster RCNN \cite{fasterrcnn}, and RetinaNet \cite{retina} on the Wildtrack dataset. We find that the multiview detection performance has the same trend as 2D detection accuracy. For example, the best 2D detection method CrowdDet also gives the highest MODA in multiview detection. These results suggest that 2D detection has a profound influence on our method. On the other hand, we compare MSPN \cite{MSPN} used in our system with Hourglass \cite{Hourglass} as Fig. \ref{DetectionBackbones}B shown. We find that MSPN with a higher standing point estimation accuracy contributes to better system performance. 
This is because correct standing point estimation plays an important role in constructing cardboard humans as the actual position on the ground plane.
\label{variant_study: Pedestrian_KeypointDetection}

{\bf Comparing different human height estimates.} In Fig. \ref{HumanFeatureHeight}B, we compare a few variants in human height estimation. ``Fixed height'' of 1.8m is used in some existing feature projection-based methods \cite{mvdet, shot, mvdetr, vfa}, which inevitably introduces noise given its inaccuracy. Moreover, we expect insufficient human description if we consider half of the body or only the foot region. These considerations are verified in this experiment, where using the whole body region found by 2D detection yields the highest MODA accuracy. Using the feet region only is the worst variant because too little appearance is integrated. 
This experiment confirms that relatively accurate height estimates are beneficial for appearance feature extraction and avoiding background noise.

\begin{figure}
\setlength{\belowcaptionskip}{-0cm}
  \centering{
    \includegraphics[width=1.0\linewidth]{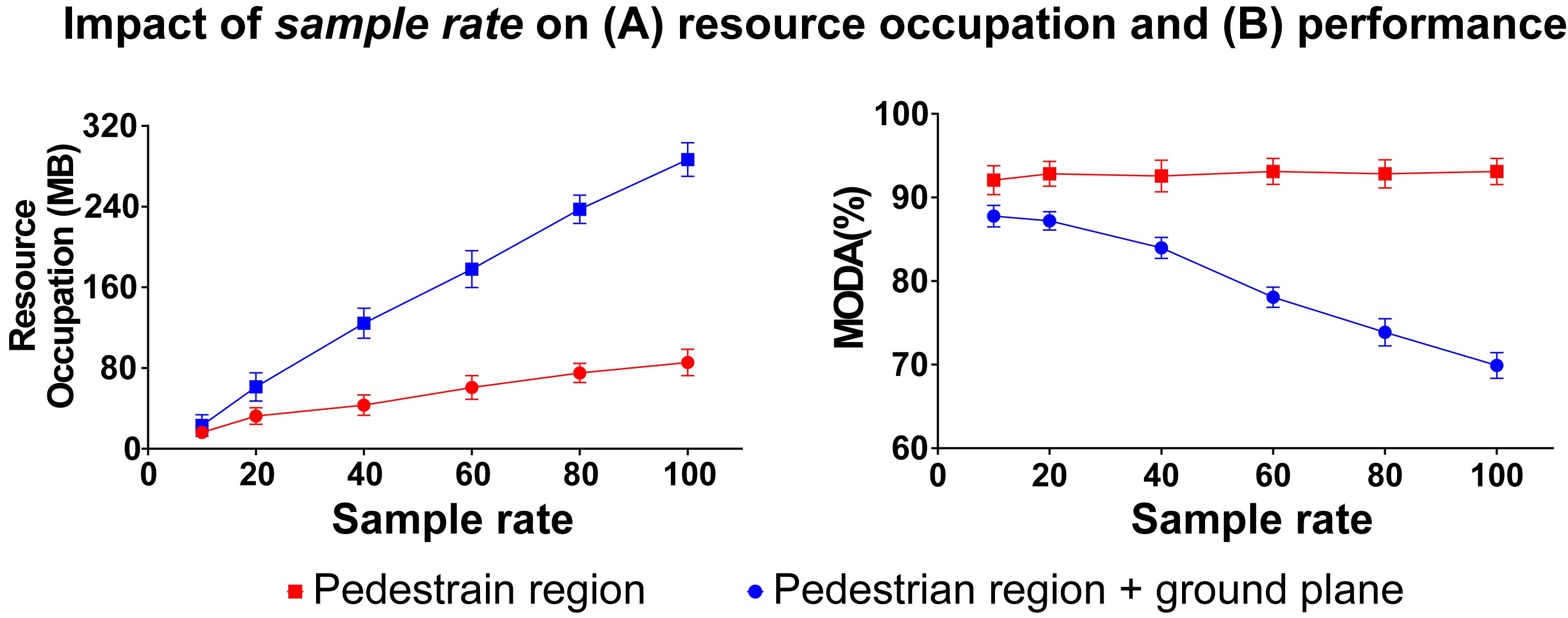}}
\caption{Comparing the performance of additional 3D modeling ground plane in our system. Two aspects are considered: (A) Storage memory consumption and (B) detection accuracy measured by MODA (\%). 
}

  \label{SampleRate}
\end{figure}

\textbf{Impact of point clouds sampling rate and ground plane 3D modeling.} 
\label{groundplane_samplerate_impact}
By default, for each cardboard human we use 50\% of its points; we also discard all the ground plane points. Here we evaluate how these two aspects (both related to point clouds) impact our system, in Fig. \ref{SampleRate}, where ``Sample rate'' means the preservation rate of the point clouds. In Fig. \ref{SampleRate}A, when we gradually increase the point clouds sampling rate, the GPU memory consumption increases linearly, and modeling the ground plane would incur additional memory costs because the ground plane itself takes up a considerable amount of memory. On the other hand, in Fig. \ref{SampleRate}B, we observe when the sample rate increases, detection accuracy remains quite stable when the ground plane is not modeled in our system; otherwise the performance drops. It is probably because the modeling of the ground plane  introduces noisy points which compromise our system. Therefore, considering both accuracy and memory consumption, we choose not to model the ground plane and to use 50\% of the points for every human point clouds. 

\begin{figure}[t]
\setlength{\abovecaptionskip}{-0cm}
  \centering{
    \includegraphics[width=0.98\linewidth]{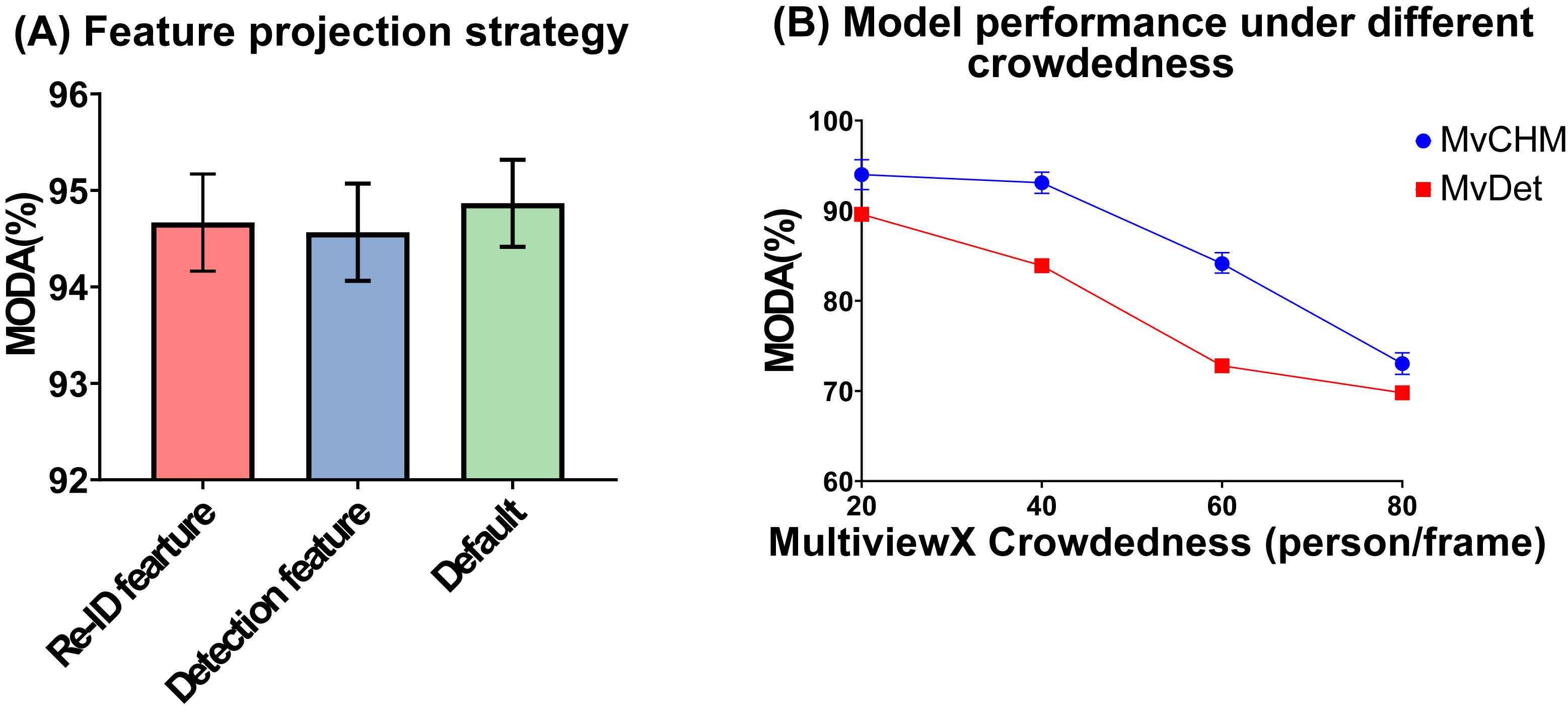}}
\caption{(A): Comparing various human description strategies with the default point clouds projection scheme. 
(B): Comparison with the feature-projection-based baseline method MVDet \cite{mvdet} under various pedestrian density levels on the MultiviewX dataset.}
  \label{ProjectionStrategies}
\end{figure}

{\bf Different human appearance descriptors in cardboard modeling.} 
By default, our method encodes human appearance features as thin cardboard-like point clouds. In this section, we explore the impact of applying different feature representation strategies on model performance. 
We compare our point cloud representation strategy with 
two other types of pedestrian features, namely, the re-ID features from an off-the-shelf person re-identification (re-ID) model \cite{OSNet} and the feature from the Feature Pyramid Network (FPN) \cite{FPN} used in our detection model \cite{CrowdDet}. Both variants follow the same training and test protocol as our system. 
Results are summarized in Fig. \ref{ProjectionStrategies}A. We observe that models with different feature representations perform similarly because all of them include similar pedestrian feature appearances to some extent. 

\phantomsection
\label{variant_study: PedestrianDensity}
{\bf Impact of pedestrian density.} In Fig. \ref{ProjectionStrategies}B, we evaluate our system under various levels of crowdedness on the MultiviewX dataset and compare it with MVDet \cite{mvdet}. We find decreased detection accuracy with increasing crowdedness, which is consistent with the findings in \cite{mvdet}. In fact, a crowded scene deteriorates the 2D detector and adds noise to the subsequent feature learning process. Furthermore, our method consistently outperforms MVDet, indicating the robustness of the proposed system. 

\section{Limitation and future work}
Our method has very competitive accuracy when it is trained and tested both on the Wildtrack and Wildtrack+ datasets. However, when tested on the MultiviewX and MultiviewX+ datasets, accuracy drops slightly (see Table \ref{PerformanceComparison}). 
Our analysis suggests the reason to be inaccurate standing point detection, causing 
the cardboard human to be poorly constructed. On the one hand, compared to the Wildtrack dataset, cameras in MultiviewX are placed lower (1.8 meters in height), which leads to the absence of pedestrian feet when they are close to the cameras. On the other hand, a high level of occlusions in MultiviewX also results in the missing of pedestrian feet. Failure cases are shown in Fig. \ref{failure} in the supplementary materials. 

There are multiple possible directions along the track of our research in future works: First, as discussed above, our method relies too much on pedestrian detection and keypoint detection. We speculate this problem can be alleviated if more accurate mapping of the human body from 2D to 3D could be established. In this regard, existing works in 3D human modeling offer a valuable source of ideas \cite{Densepose, smpl}. 
Second, instead of modeling the scene with explicit point cloud representation, it is possible to model the entire 3D space with implicit representation (NeRF-base methods \cite{nerf}). This paper offers a brand-new insight that the coarse but correct reconstruction of scenes can effectively integrate multiview clues and accurately locate targets. We hope that our findings will motivate the progress of multiview detection.

\section{Conclusion}
Under the context of multiview pedestrian detection, 
this paper proposes a new pedestrian representation that models the human as one-channel point clouds, much like standing cardboard. This modeling method results from a reasonable use of the scene geometry and allows for the effective fusion of pedestrian position and appearance features through point cloud feature learning. Moreover, because pedestrians are explicitly separated from each other and the background, less noise is included compared with feature project-based methods. Our system is evaluated on two existing multiview detection datasets and their extension datasets where we report very competitive detection accuracy compared with the state-of-the-art methods.

{\small
\bibliographystyle{ieee_fullname}
\bibliography{MvCHM}
}

\newpage\clearpage


\appendix

\twocolumn[
\centering
\Large
\textbf{Multiview Detection with Cardboard Human Modeling} \\
\vspace{0.5em}Supplementary Material \\
{%
\renewcommand\twocolumn[1][]{#1}%
\maketitle

\begin{center}
    \begin{minipage}{0.7\linewidth}
    \vspace{3pt}
    \centerline{\includegraphics[width=\textwidth]{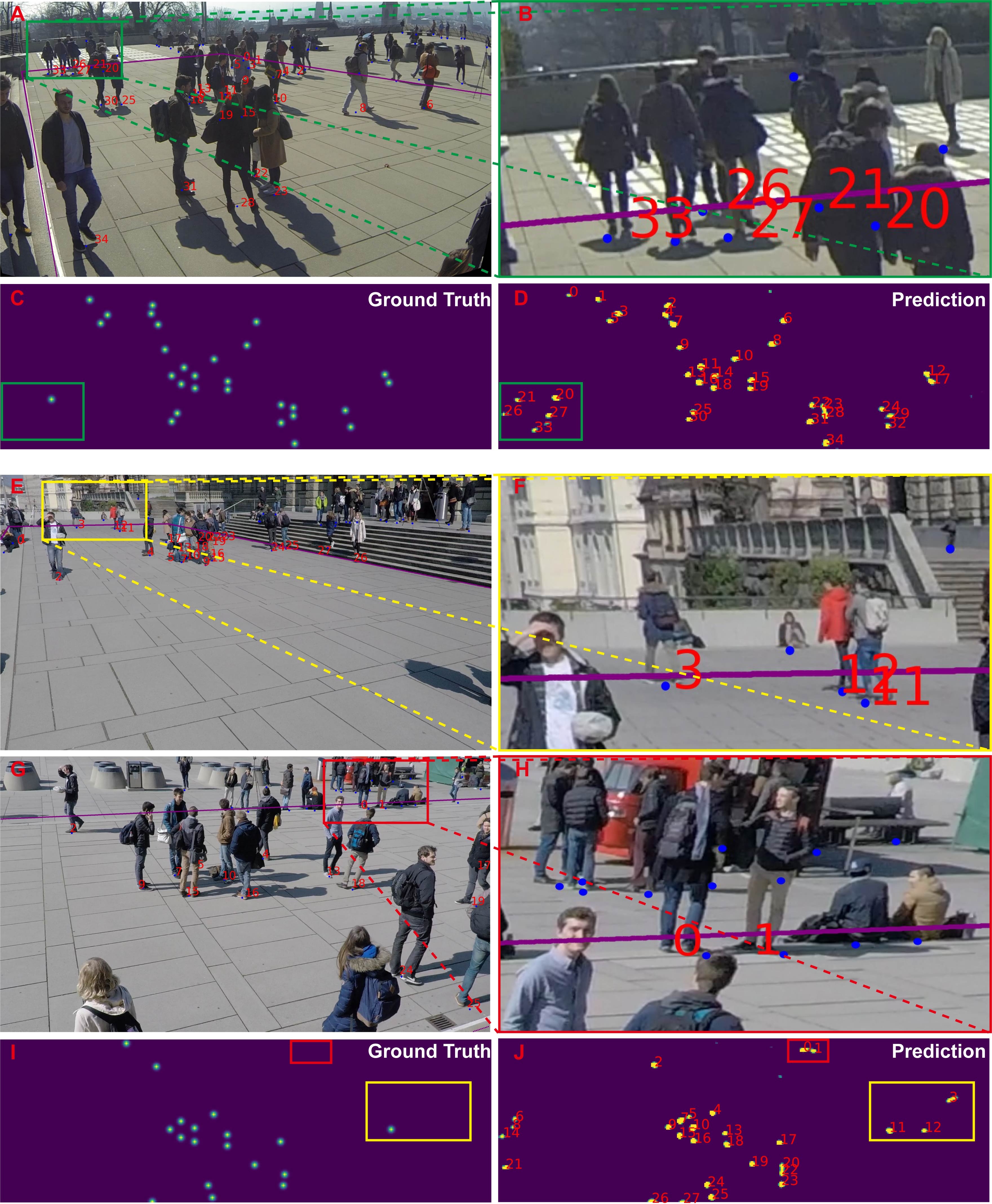}}
    \end{minipage}
\end{center}%
\vspace{-1em}
\captionof{figure}{Examples of the missing pedestrian annotation in the Wildtrack dataset. A-C figures are the data with subscript ``00001975.png'', and E-I figures are the data with subscript ``00001805.png''.}
\vspace{0.5em}
\label{label_omission_img}
}] %

\section{Appendix}


\subsection{Discussion on the missing annotations}
\begin{figure*}[t]
    \centering
    \includegraphics[width=0.72\linewidth]{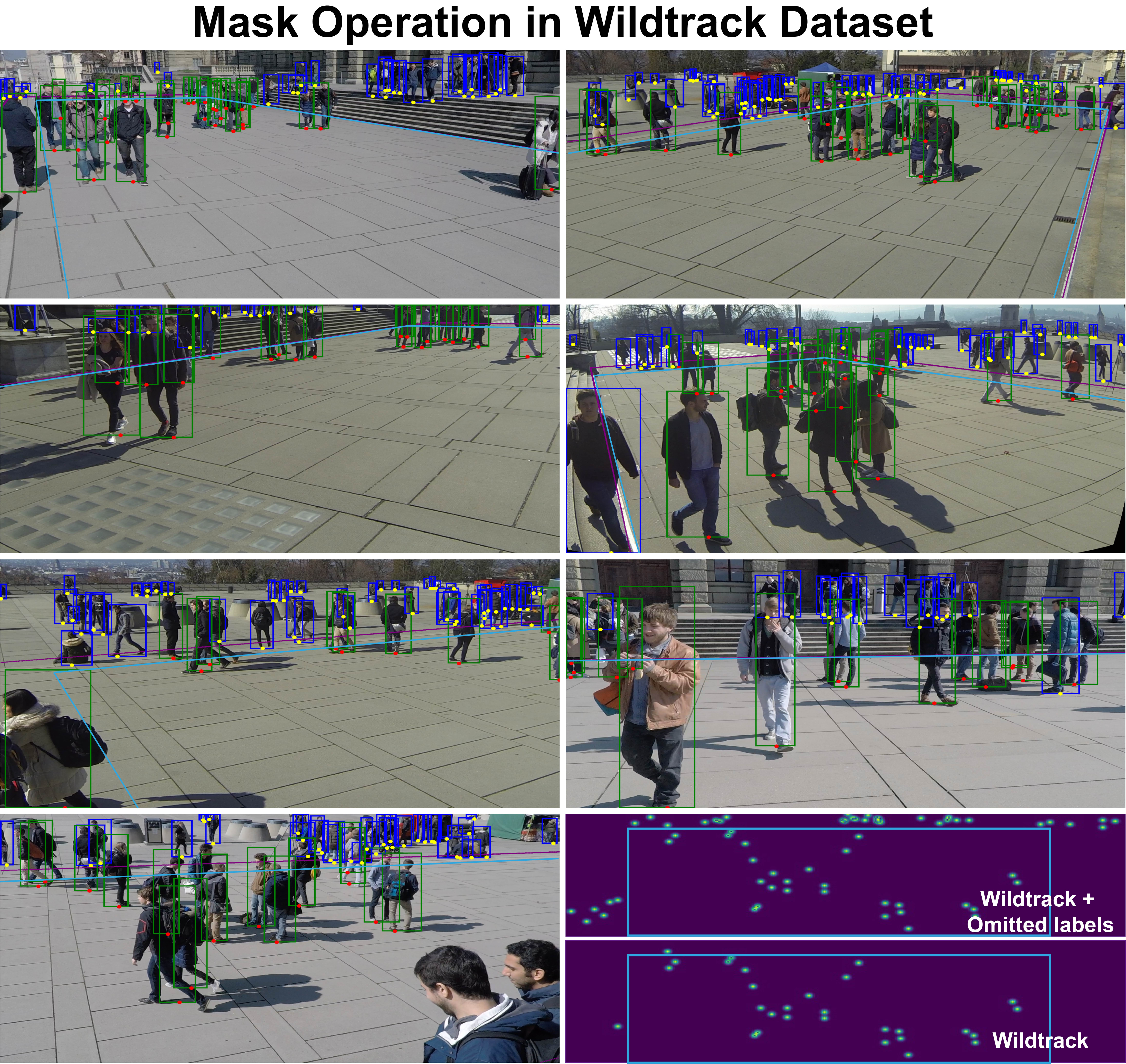}
    \caption{The visualization of the masking operation applied on the Wildtrack dataset. In each camera view, the origin detection area defined in the original dataset is bounded with \purple{purple lines}, while the area boundaries after masking are colored in \blue{blue}. As illustrated in the BEV heatmap, we identify an enormous amount of missing annotations near the edge of the detection plane (lies in between the \purple{purple} ground area and the \blue{blue} masked area), the mask is applied to filter most of the missing labels and ambiguities. The masked detection area is marked by bright blue color in the last two BEV heatmaps. }
    \label{WildtrackMask}
\end{figure*}

\begin{figure*}[h]
    \begin{minipage}{0.34\linewidth}
    \vspace{3pt}
    \centerline{\includegraphics[width=\textwidth]{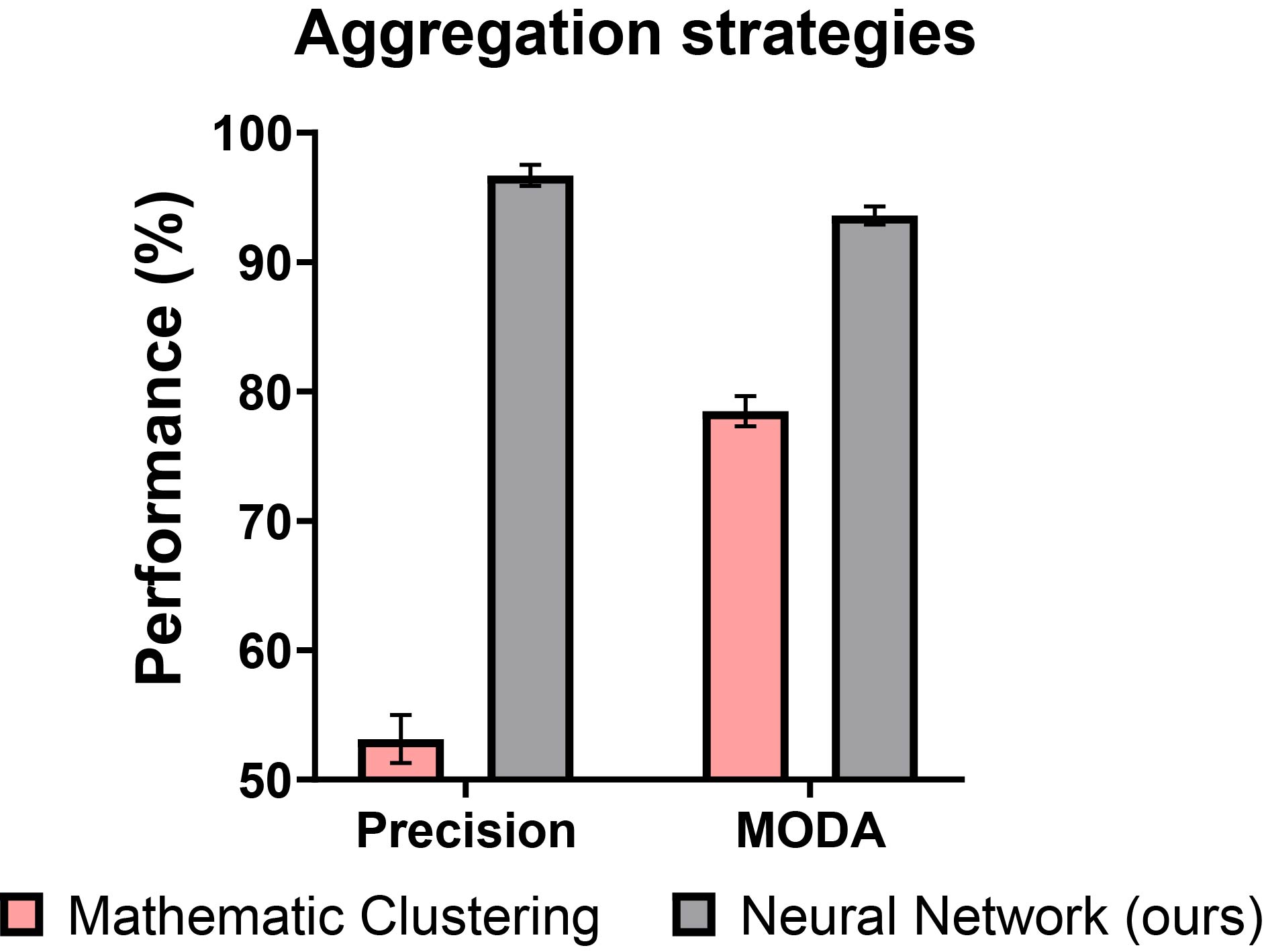}}
    \centerline{{\bf A}}
    \end{minipage}
    \begin{minipage}{0.6\linewidth}
    \vspace{3pt}
    \centerline{\includegraphics[width=\textwidth]{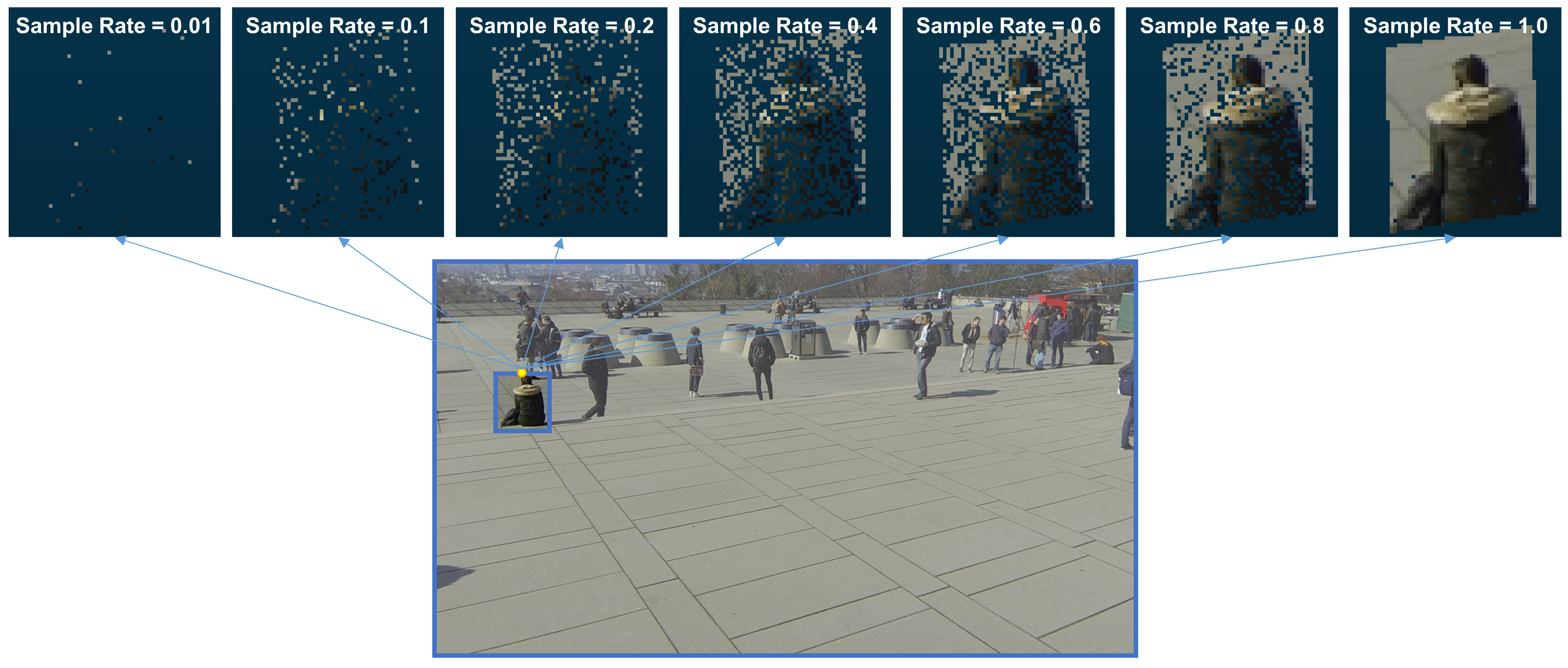}}
    \centerline{{\bf B}}
    \end{minipage}
    \captionof{figure}{(A) Comparing different aggregation strategies after projection. (B) The visualization of different point clouds sampling rates.}
    \label{appendix_aggregation_rate}
\end{figure*}

\label{Label_Ommision}
As mentioned in Section \ref{sota}, we identify severe label omissions in the Wildtrack dataset, two typical examples are shown in Fig. \ref{label_omission_img}. In the first image batch marked with green boxes, 4 persons in the bottom left corner are neglected, however, our algorithm successfully predicts these persons' locations. In the other batch of images marked with red and yellow boxes, our system demonstrates that there are more people neglected near the edge of the detection area. According to the image, these persons are standing inside the detection area bounded by the purple lines, while their labels are not provided. The lack of such annotations may cause a \textit{`fake'} high false positive rate if the algorithm successfully makes the prediction on that target. Therefore, for a fair comparison on the Wildtrack dataset, we apply a mask on the predicted BEV map for all the previous methods, specifically, we define an area where the label omission has a high occurrence rate (usually refers to the area that nears the edge of the square), and we simply ignore the prediction results in this area to avoid \textit{`fake'} high false positive rate, the shape of the mask is shown in Fig. \ref{WildtrackMask}.

\begin{figure}
    \centerline{\includegraphics[width=.49\textwidth]{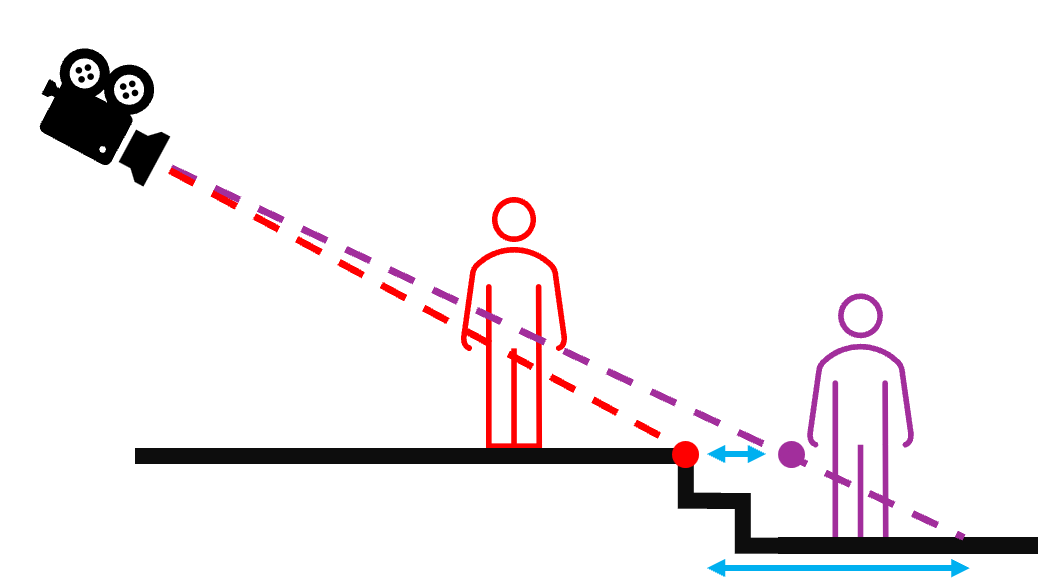}}
    \captionof{figure}{A side view of the detection plane and the stairs occurred at its edge shown in Fig. \ref{label_omission_img}H. This demonstrates how trivial calibration error leads to huge edge shifting of the detection plane in the Wildtrack dataset. The red point represents the actual edge of the detection plane, and the purple point stands for the erroneous edge calculated according to the camera calibration file. }
    \label{edge}
\end{figure}

\begin{figure}[h]
    \setlength{\belowcaptionskip}{-0.3cm}
    \centerline{\includegraphics[width=.4\textwidth]{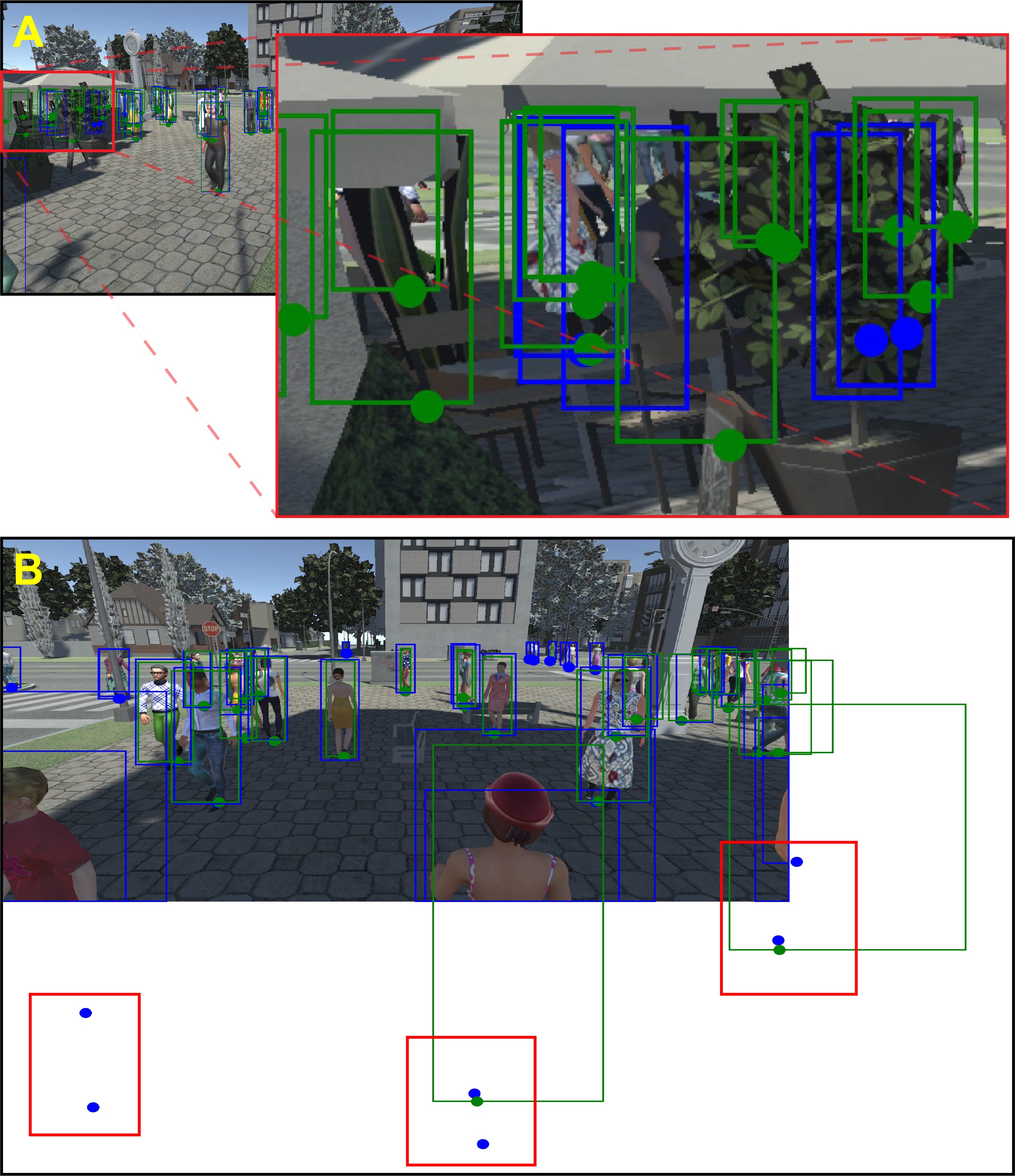}}
    \captionof{figure}{Failure cases on the MultiviewX dataset. Green boxes and dot points are ground truth and blue boxes and dot points are detection results. Failure detections are caused by (A) severe occlusion and (B) the absence of pedestrians' feet.}
    \label{failure}
\end{figure}

We further analyze the source of these unlabelled targets. As shown in the BEV map of Fig. \ref{WildtrackMask}, despite few errors made by human annotators, most of the ambiguities occur around the top edge of the detection square due to trivial camera calibration inaccuracy and the special architectural structure of the square, i.e., the stairs occurred at the edge of the square magnify the calibration error. As demonstrated in Fig. \ref{label_omission_img}H, the defined detection plane edge colored in purple is slightly shifted from the actual edge of the plane, and coincidentally, a stair occurs at the edge of the square, which drastically enlarges the disparity between the actual edge and the shifted one by adding an extra distance in $Z$ axis, as demonstrated in Fig. \ref{edge}. In this figure, the person (or the person's lower leg) colored in purple is counted as a target standing inside the detection plane in our proposed system when seeing from particular views. In fact, all previous methods that utilize camera calibrations suffer from this problem to a specific extent on the Wildtrack dataset. While with the mask, these ambiguities are avoided.

We argue that the evaluation process remains valid with this mask since the mask only covers the areas near the edge of the detection area where the crowdedness and occlusion of the pedestrians are relatively low compared with the center area. Thus, the performance of the algorithms on localizing targets under crowdedness and occlusion can be evaluated as equally as on the origin Wildtrack dataset. All the previous methods report higher performance scores after masking.

%


\subsection{Different point clouds sampling rate of ROI}
\begin{centering}
To clearly demonstrate the sample rate shown in Fig. \ref{SampleRate}, we visualize the results of one of the detection regions in 3D space. As Fig. \ref{appendix_aggregation_rate}B shows, the first row of the figure shows that we take random sampling operation. The sample rate between 0.4 and 0.6 not only represents sufficient appearance features of the target but also reduces the storage of the point clouds to a certain extent. In our proposed pipeline, we set 0.5 as the default sample rate.
\label{Sampling}
\end{centering}
\subsection{Benefit of neural network for aggregation}

To highlight the significance of our neural network-based point clouds aggregation procedure, we compare our method with the clustering pipeline proposed in \cite{Lima_2021_CVPR}. To ensure the fairness of the experiment, we use the standing point as the position feature and the high-dimensional re-ID feature as the appearance feature. We cluster these high-dimensional features using the same clique-based clustering method introduced in \cite{Lima_2021_CVPR}. The performance is shown in Fig. \ref{appendix_aggregation_rate}(A) well illustrates the efficiency of adopting neural networks to aggregate point clouds.


\subsection{Depth estimation using ray tracing}
\label{ray-tracing-derivation}
Depth of localized ROI is essential for modeling cardboard humans, due to the lack of depth value labels, we adopt the ray-tracing (\cite{ray-tracing}) technique to calculate the depth for each ROI localization result, namely each bounding box region. For each pedestrian detection result, we calculate the depth of the head and estimated standing point, and fill the rest of the area with interpolated depth values. With the calculated depth, we can project the 2D ROI localization results back into the 3D space to form 3D cardboards.

Given ray-tracing formula \cref{ray-tracing-formula}, we define the standing point as $P_\text{standpoint} = [P_\text{x}^\text{s}, P_\text{y}^\text{s}, P_\text{z}^\text{s}]$, and head point $P_\text{head} = [P_\text{x}^\text{h}, P_\text{y}^\text{h},P_\text{z}^\text{h}]$. 

\subsubsection{Depth of the standing point}
We first calculate the 3D coordinate of the pedestrian standing point. For each standing point, we further define the camera 3D center as $O = [O_\text{x}, O_\text{y}, O_\text{z}]$, the direction of the ray direction $D_\text{standpoint}$ from the camera center to the standing point as $D_\text{standpoint} = [D_\text{x}^\text{s}, D_\text{y}^\text{s}, D_\text{z}^\text{s}]$, and the distance between the camera center and standing point on the object as $t$. The ray tracing formula is denoted as:
\begin{equation}
\begin{bmatrix}
    P_\text{x}^\text{s} \\
    P_\text{y}^\text{s} \\
    P_\text{z}^\text{s}
\end{bmatrix}
 = 
 \begin{bmatrix}
    O_\text{x} \\
    O_\text{y} \\
    O_\text{z}
\end{bmatrix}
+ t
\begin{bmatrix}
    D_\text{x}^\text{s}\\
    D_\text{y}^\text{s}\\
    D_\text{z}^\text{s}
\end{bmatrix}
\text{   i.e.}
\begin{cases}
  P_\text{x}^\text{s} = O_\text{x} + tD_\text{x}^\text{s}  \\
  P_\text{y}^\text{s} = O_\text{y} + tD_\text{y}^\text{s}  \\
  P_\text{z}^\text{s} = O_\text{z} + tD_\text{z}^\text{s}  \\
\end{cases}
\label{standing_point_base_formula}
\end{equation}

\label{Ray-tracing-appendix}
Given the premise that the standing point is on the ground plane, where $Z = 0$, we have $P_\text{z}^\text{s} = 0$:
\begin{equation}
  O_\text{z} + tD_\text{z}^\text{s} = 0
\end{equation}
hence,
\begin{equation}
  t = - \frac{O_\text{z}}{D_\text{z}^\text{s}}
  \label{get_t}
\end{equation}
substitute $t$ into \cref{standing_point_base_formula}:
\begin{equation}
  \begin{cases}
    P_\text{x}^\text{s} = O_\text{x} - \frac{O_z}{D_z}D_\text{x}^\text{s}  \\
    P_\text{y}^\text{s} = O_\text{y} - \frac{O_z}{D_z}D_\text{y}^\text{s}  \\
    P_\text{z}^\text{s} = 0
  \end{cases}
  \label{standing_point_sub}
\end{equation}
Now, to determine $P_\text{x}^\text{s}$ and $P_\text{y}^\text{s}$, we need to further explore the camera position $O$ and ray direction $D_\text{standpoint}$. 
The camera position $O$ in the world coordinate system is determined with:
\begin{align}
  O = -R^TT
\end{align}
where $R$ and $T$ are the rotation matrix and translation matrix that map the object from the world coordinate to camera coordinate, and $-R^TT$ is a $3\times1$ matrix.

Next, to get the direction from the camera center to the standing point $D_\text{standpoint}$ of the ray, we need to find the correlation between the ray and the world coordinates systems. We first determine the ray direction inside the camera, which is to define the ray that starts from the camera origin to the pixel coordination system, and furthermore, we translate the ray from the pixel coordinate system to the camera coordinates system using the intrinsic matrix, and finally, we project the origin-to-camera ray to an origin-to-world one. Assume the standing point in the pixel coordinate system is marked as $[u^\text{s}, v^\text{s}]$ and the camera has the intrinsic matrix $k$ as:
\begin{equation}
    k = 
    \begin{bmatrix}
        fx& 0& cx\\
        0& fy& cy\\
        0& 0& 1
    \end{bmatrix}
\end{equation}
where $fx, fy$ represent the focal length in $x, y$ direction, $cx, cy$ are the translation between camera coordinates systems and pixel coordinates systems. We could now define the standing point in the camera coordinate system $[X_\text{cam}^\text{s}, Y_\text{cam}^\text{s}, Z_\text{cam}^\text{s}]$ with the following derivation:
\begin{align}
Z_\text{cam}^\text{s}
\begin{bmatrix}
    u^\text{s} \\
    v^\text{s} \\
    1
\end{bmatrix}
& = [K | 0]
\begin{bmatrix}
    X_\text{cam}^\text{s} \\
    Y_\text{cam}^\text{s} \\
    Z_\text{cam}^\text{s} \\
    1
\end{bmatrix}\\
&= 
\begin{bmatrix}
    fx& 0& cx & 0\\
    0& fy& cy & 0\\
    0& 0& 1 & 0
\end{bmatrix}
\begin{bmatrix}
    X_\text{cam}^\text{s} \\
    Y_\text{cam}^\text{s} \\
    Z_\text{cam}^\text{s} \\
    1
\end{bmatrix}
\end{align}

\begin{align}
& \Rightarrow
    \begin{cases}
        X_\text{cam}^\text{s}u^\text{s} = X_\text{cam}^\text{s}fx + Z_\text{cam}^\text{s}cx \\
        Y_\text{cam}^\text{s}v^\text{s} = Y_\text{cam}^\text{s}fy \ + Z_\text{cam}^\text{s}cy \\
        Z_\text{cam}^\text{s} = Z_\text{cam}^\text{s}
    \end{cases} \\ 
    &\Rightarrow
    \begin{cases}
        X_{cam}^\text{s} = \frac{Z_{cam}^\text{s}(u^\text{s} - cx)}{fx}  \\
        Y_{cam}^\text{s} = \frac{Z_{cam}^\text{s}(v^\text{s} - cy)}{fy}  \\
        Z_\text{cam}^\text{s} = Z_\text{cam}^\text{s}
    \end{cases}
  \label{XY_cam_formula}
\end{align}

We observe that $Z_\text{cam}^\text{s}$ is still unknown. However, since the calculation target is the ray direction, which is not affected by the length of the ray, we divide the $Z_\text{cam}^\text{s}$ in each line on the right of the equations to obtain the normalized origin-to-camera direction $D_\text{o2c}$, denoted as:
\begin{equation}
    D_\text{o2c} = 
    \begin{bmatrix}
        X_{cam}^\text{s}\\
        Y_{cam}^\text{s}\\
        Z_{cam}^\text{s}
    \end{bmatrix}
     = 
    \begin{bmatrix}
        \frac{u^\text{s} - cx}{fx}\\
        \frac{v^\text{s} - cy}{fy}\\
        1
    \end{bmatrix}
    \label{Do2c_formula}
\end{equation} 
Finally, we project the origin-to-camera ray direction to the origin-to-world direction using the inverse of rotation matrix $M$ (\cite{nerf}). Therefore the final origin-to-world direction $D_\text{standpoint}$ is represented as:

\begin{equation}
    D_\text{standpoint} =
    \begin{bmatrix}
        D_\text{x}^{s}\\
        D_\text{y}^{s}\\
        D_\text{z}^{s}
    \end{bmatrix}
    = D_{o2c} \cdot M^{-1} =
    \begin{bmatrix}
        \frac{u^\text{s} - cx}{fx}\\
        \frac{v^\text{s} - cy}{fy}\\
        1
    \end{bmatrix}  \cdot M^{-1}
    \label{Dstand_formula}
\end{equation}
With $O, D_\text{standpoint}$, $P_\text{z}^\text{s}, P_\text{y}^\text{s}$ in \cref{standing_point_base_formula} can be determined. We now know the exact 3D world coordinate $P_\text{standpoint} = [P_\text{x}^s, P_\text{y}^s, P_\text{z}^s]^\text{T}$ of the standing point. Lastly, we project the standing point $P_\text{standpoint}$ to the camera coordinate system using the extrinsic matrix to obtain $[X_{cam}^\text{s}, Y_{cam}^\text{s}, Z_{cam}^\text{s}]$ in which $Z_\text{cam}^\text{s}$ is not divided, and the $Z_\text{cam}^\text{s}$ value is the \textit{depth} of the standing point.

\subsubsection{Depth of the head point}
To calculate the 3D world coordinate of the head, we leverage the assumption that the head and standing point of the same pedestrian lie on the same vertical line, therefore, both head and standing point share the same $P_\text{x}$ and $P_\text{y}$. The actual height in the real world is the calculation target, and we regard the top of each bounding box as the head of the pedestrian in each 2D image. Therefore. Recall the definition in \cref{ray-tracing-derivation}, we have:

\begin{equation}
\begin{bmatrix}
    P_\text{x}^\text{s} \\
    P_\text{y}^\text{s} \\
    P_\text{z}^\text{s}
\end{bmatrix}
 = 
 \begin{bmatrix}
    O_\text{x} \\
    O_\text{y} \\
    O_\text{z}
\end{bmatrix}
+ t
\begin{bmatrix}
    D_\text{x}^\text{h}\\
    D_\text{y}^\text{h}\\
    D_\text{z}^\text{h}
\end{bmatrix}
\text{   i.e.}
\begin{cases}
  P_\text{x}^\text{h} = O_\text{x} + tD_\text{x}^\text{h}  \\
  P_\text{y}^\text{h} = O_\text{y} + tD_\text{y}^\text{h}  \\
  P_\text{z}^\text{h} = O_\text{z} + tD_\text{z}^\text{h}  \\
\end{cases}
\label{head_base_formula}
\end{equation}

In this case, $P_\text{x}^\text{h}, O_\text{x}, O_\text{y}$ are known and $D_\text{x}^\text{h}$ can be calculated according to \cref{XY_cam_formula} - \cref{Dstand_formula}. Thus, the only unknown $t$ can be calculated by substituting $P_\text{x}^\text{h} = P_\text{x}^\text{s}$ into \cref{get_t}. Hence, with all the calculated variables, $P_\text{z}^\text{h}$ is obtained. The depth of the head is then acquired following the same steps as in the depth calculation for the standing point.

\subsection{Point clouds generation for each ROI area}
We fill each localized ROI region with interpolated depth values according to the calculated depth of the standing point and head. For each pixel in the localized ROI area, the 3D coordinate(point cloud) $P_\text{all} = [P_\text{x}^\text{a}, P_\text{x}^\text{a}, P_\text{x}^\text{a}]$ is obtained with the following formula:

\begin{equation}
\begin{bmatrix}
    P_\text{x}^\text{all} \\
    P_\text{y}^\text{all} \\
    P_\text{z}^\text{all} \\ 
    1
\end{bmatrix}
 = 
\begin{bmatrix}
    R & t \\
    0 & 1
\end{bmatrix}_{4\times4}^\text{-1}
\cdot 
\begin{bmatrix}
    X_{cam}^\text{all}\\
    Y_{cam}^\text{all}\\
    Z_{cam}^\text{all}\\
    1
\end{bmatrix}
\end{equation}
Where 
\begin{equation}
\begin{cases}
        X_{cam}^\text{all} = \frac{Z_{cam}^\text{all}(u^\text{all} - cx)}{fx}  \\
        Y_{cam}^\text{all} = \frac{Z_{cam}^\text{all}(v^\text{all} - cy)}{fy}  \\
        Z_\text{cam}^\text{all} = \text{interpolated depth}
\end{cases}
\label{head_base_formula}
\end{equation}
$R, t$ is the rotation and translation matrix of the camera.

We now have the 3D world coordinates(point clouds) for every pixel in a localized ROI area.

\begin{figure*}[t]
    \centering
    \includegraphics[width=0.72\linewidth]{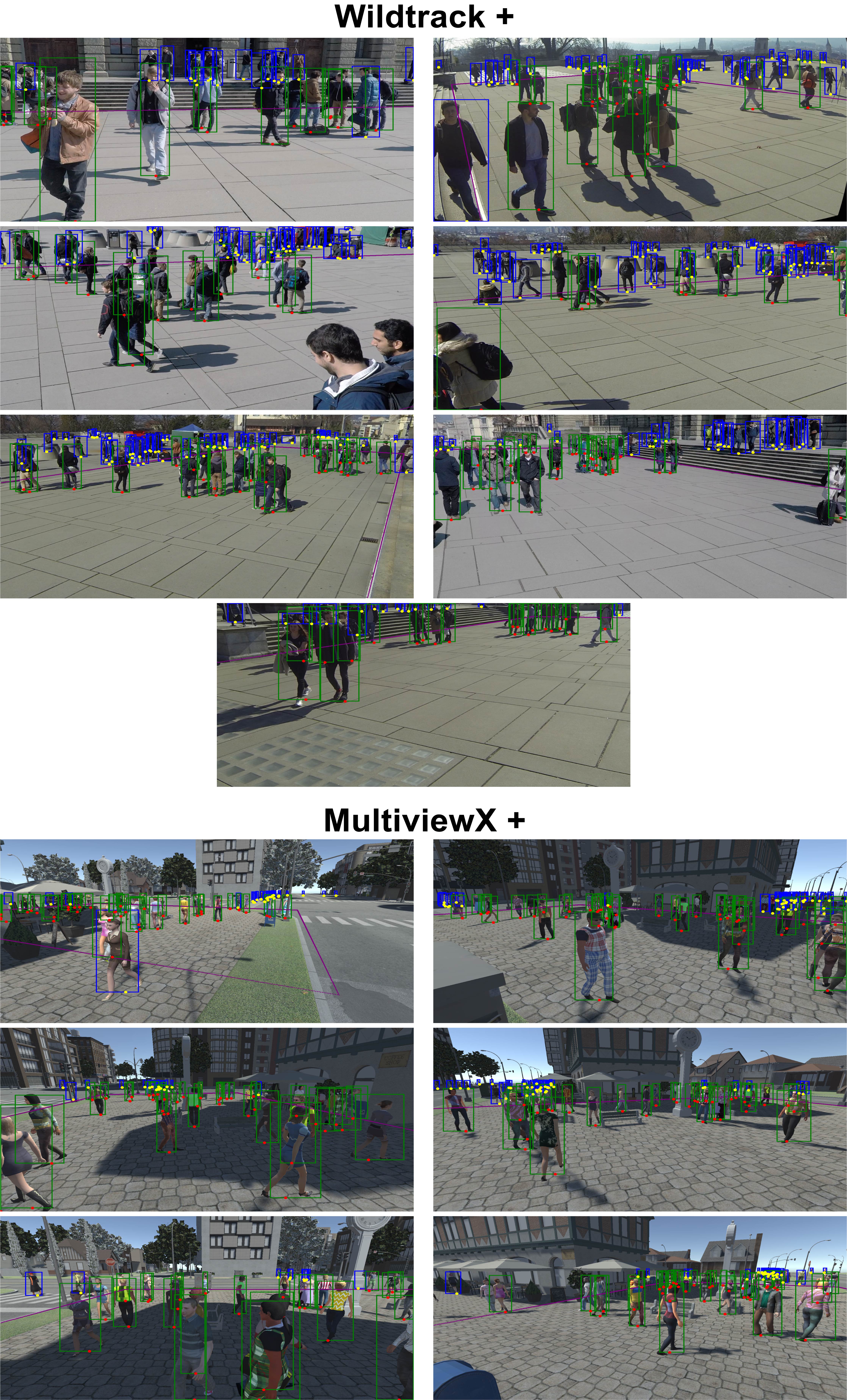}
    \caption{Label visualization of the Wildtrack+ and MultiviewX+ dataset. For official Wildtrack and MultiviewX, we use \green{green bounding box} and \red{red dot point} to visualize the region of interest (ROI). For the Wildtrack+ and MultiviewX+ we proposed, we additionally annotate the pedestrians outside the detection area (bounded by \purple{purple lines}). The supplementary labels are painted as \blue{blue bounding boxes} and {\color{light-yellow} yellow dot point}. }
    \label{Wildtrack+_MultiviewX+}
\end{figure*}
\label{appendix}

\begin{figure*}[t]
    \centering
    \includegraphics[width=0.72\linewidth]{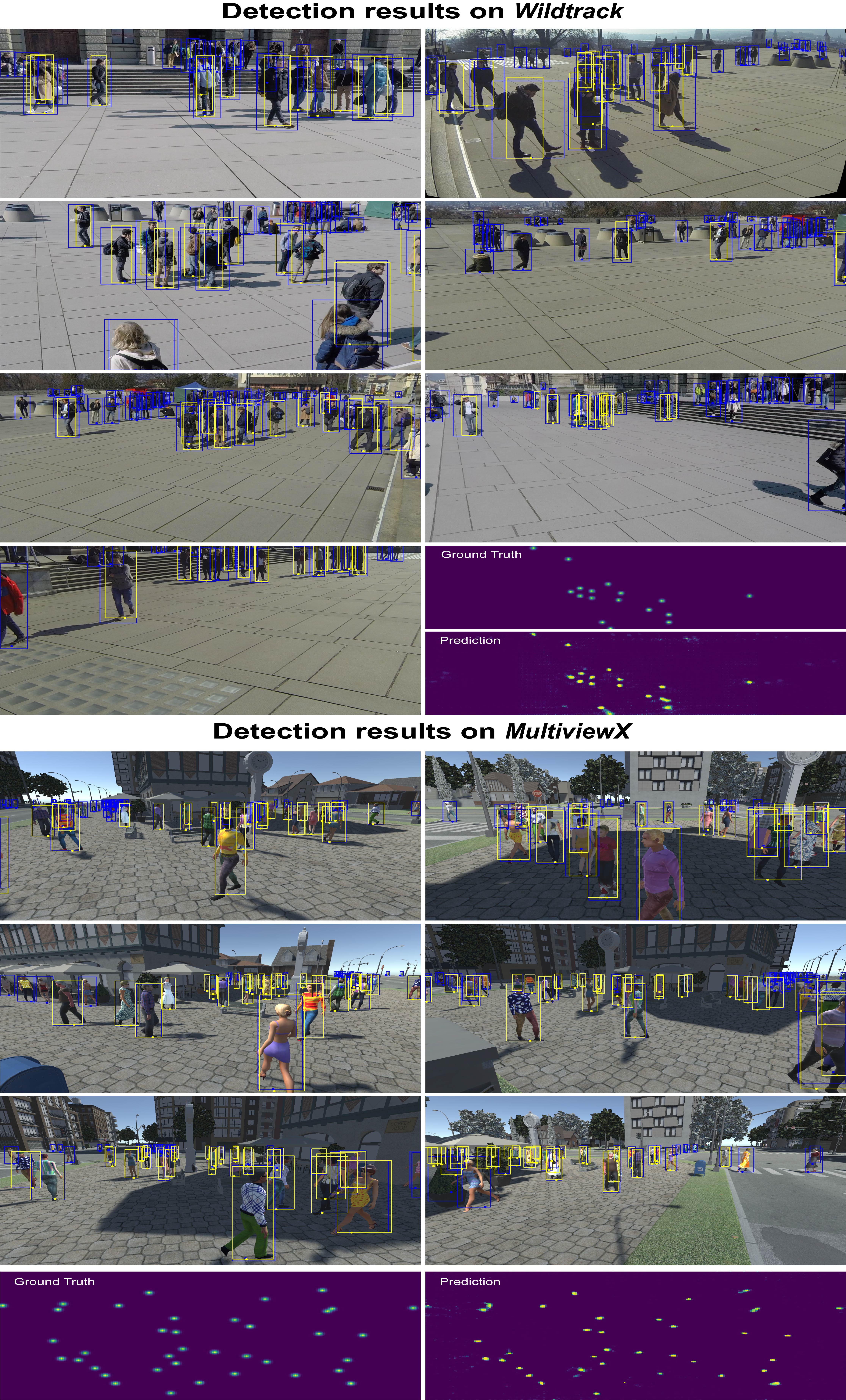}
    \caption{Detection results on Wildtrack and MultiviewX dataset.  Ground truth including standing points and bounding boxes are labeled by {\color{light-yellow} yellow color}. And the detection results are labeled by \blue{blue color}.}
    \label{DetectionResults}
\end{figure*}

\end{document}